\documentclass[10pt,twocolumn,letterpaper]{article}

\usepackage[pagenumbers]{cvpr} % To force page numbers, e.g. for an arXiv version

\definecolor{cvprblue}{rgb}{0.21,0.49,0.74}
\usepackage[pagebackref,breaklinks,colorlinks,allcolors=cvprblue]{hyperref}

\def\paperID{2616} % *** Enter the Paper ID here
\def\confName{CVPR}
\def\confYear{2025}

\title{CoMM: A Coherent Interleaved Image-Text Dataset \\ for Multimodal Understanding and Generation}

\author{
  {Wei Chen}$^{1, 2}$\thanks{Equal contribution. Work was done when Wei Chen interned in Kuaishou. $^{\dagger}$Corresponding authors.},
  \quad
  {Lin Li}$^{1,3*}$,
  \quad
  {Yongqi Yang}$^{2*}$, \\
  \quad
  {Bin Wen$^4$,}
  \quad
  {Fan Yang$^4$,}
  \quad
  {Tingting Gao$^4$,}
  \quad
  {Yu Wu$^{2\dagger}$}\textbf{,}
  \quad
  {Long Chen}$^{1\dagger}$ \\
  $^1$The Hong Kong University of Science and Technology,
  $^2$Wuhan University, \\
  $^3$AI Chip Center for Emerging Smart Systems, 
  $^4$Kuaishou Technology\\ % AI Chip Center for Emerging Smart Systems
  {\tt\small {\{wchendb, lllidy, longchen\}@ust.hk, \{yongqiyang, wuyucs\}@whu.edu.cn}} \\
  {\small\url{https://github.com/HKUST-LongGroup/CoMM}} \\
}

\newcommand{\datasetname}{CoMM}
\usepackage[utf8]{inputenc} % allow utf-8 input
\usepackage[T1]{fontenc}    % use 8-bit T1 fonts
\usepackage{hyperref}       % hyperlinks
\usepackage{url}            % simple URL typesetting
\usepackage{booktabs}       % professional-quality tables
\usepackage{amsfonts}       % blackboard math symbols
\usepackage{nicefrac}       % compact symbols for 1/2, etc.
\usepackage{microtype}      % microtypography
\usepackage{xcolor}         % colors
\usepackage{multirow} 
\usepackage{graphicx}
\usepackage{amsmath} 
\usepackage{wrapfig}
\usepackage{caption}
\usepackage{subcaption}
\usepackage{enumerate}
\usepackage{enumitem}
\usepackage{tcolorbox}
\usepackage{verbatim}
\usepackage{stfloats} 
\usepackage{arydshln}

\usepackage{listings}
\usepackage{color}
\definecolor{dkgreen}{rgb}{0,0.6,0}
\definecolor{gray}{rgb}{0.5,0.5,0.5}
\definecolor{mauve}{rgb}{0.58,0,0.82}
\begin{document}
\maketitle
\begin{abstract}  
    Interleaved image-text generation has emerged as a vital multimodal task aimed at creating sequences of interleaved visual and textual content given a query. Despite notable advancements in recent multimodal large language models (MLLMs), generating integrated image-text sequences that exhibit \emph{narrative coherence} and \emph{entity and style consistency} remains challenging due to poor training data quality. To this end, we introduce \textbf{\datasetname}, a high-quality \textbf{Co}herent interleaved image-text \textbf{M}ulti\textbf{M}odal dataset designed to enhance the coherence, consistency, and alignment of generated multimodal content. Initially, \datasetname\ harnesses raw data from diverse sources, focusing on instructional content and visual storytelling, establishing a foundation for coherent and consistent content. To further refine the data quality, we devise a multi-perspective filter strategy that leverages advanced pre-trained models to ensure the \emph{development of sentences}, \emph{consistency of inserted images}, and \emph{semantic alignment between them}. Various quality evaluation metrics are designed to prove the high quality of the filtered dataset. Meanwhile, extensive few-shot experiments on various downstream tasks demonstrate \datasetname's effectiveness in significantly enhancing the in-context learning capabilities of MLLMs. Moreover, we propose four new tasks to evaluate MLLMs' interleaved generation abilities, supported by a comprehensive evaluation framework. We believe \datasetname\ opens a new avenue for advanced MLLMs with superior multimodal in-context learning and understanding ability.

\end{abstract}    

\section{Introduction}

Interleaved image-text generation has burgeoned as an emerging multimodal task, striving to imitate the human-like capability to alternately create visual and textual content~\cite{an2023openleaf, zhu2024multimodal}. Specifically, it aims to generate a sequence of interleaved text descriptions and illustrative images given a query~\cite{an2023openleaf}. With its interleaved multimodal generation ability, state-of-the-art models can facilitate a variety of applications, \eg, multimodal instruction generation, tutorial step generation, and visual storytelling. 

\begin{figure*}
    \centering
    \includegraphics[width=1.0\linewidth]{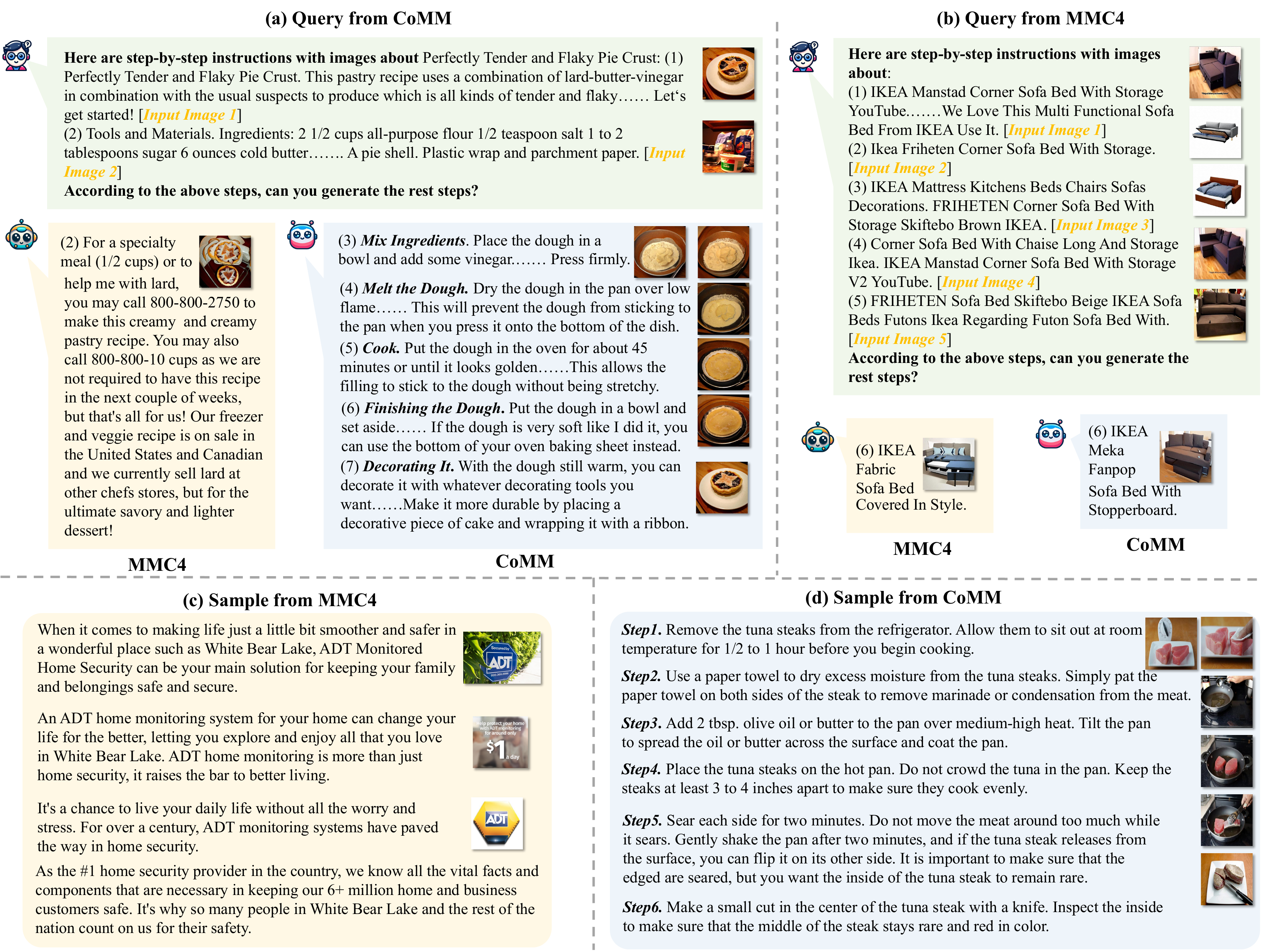}
    \vspace{-2em}
    \caption{Illustration of interleaved image-text content generation results and dataset quality. (a) Given the query from \datasetname, the interleaved image-text content generation results from the model Emu2~\cite{sun2023generative} separately trained by MMC4~\includegraphics[scale=0.2]{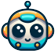}~\cite{zhu2024multimodal} and \datasetname~\includegraphics[scale=0.2]{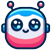} (\textbf{Ours}). (b) The query is from the MMC4. (c) A training sample is from the MMC4. (d) A training sample is from the our \datasetname.}
    \label{fig:intro_generation_results_comparison}
    \vspace{-1em}
\end{figure*}

Recent advancements in multimodal large language models (MLLMs) have exhibited exceptional capabilities in cross-modal generation, but they still struggle to generate interleaved image-text sequences coherently~\cite{sun2023generative,zheng2023minigpt}. The main reason is that most of them are trained on single image-text pairs, limited in generating coherent and contextually integrated multimodal content~\cite{zheng2023minigpt,koh2024generating}. To this end, some efforts integrate multiple models (\eg, GPT-4~\cite{achiam2023gpt} and Stable Diffusion XL~\cite{podell2023sdxl}) to generate open-domain interleaved image-text contents with arbitrary formats in a \textit{training-free} manner~\cite{an2023openleaf}. However, their results are often dominated by the generated text descriptions, limited by the generation model's understanding of these textual inputs. 
% \textcolor{red}{In contrast, other researches concentrate on constructing interleaved image-text datasets aimed at training models with both seamless multimodal understanding and generation ability (\eg, MMC4~\cite{zhu2024multimodal})}.
In contrast, other research efforts~\cite{dong2023dreamllm, ge2024seed} have attempted to train models with seamless multimodal understanding and generation capabilities using existing interleaved image-text datasets (\eg, MMC4~\cite{zhu2024multimodal}).

Albeit the unprecedented progress, it is becoming increasingly evident that data-centric methods still suffer several limitations: 1) \textbf{Narrative Coherence}: The contextual relevance and coherence of generated image-text sentences are relatively low. In Figure~\ref{fig:intro_generation_results_comparison}(a), the model pre-trained on the MMC4~\cite{zhu2024multimodal} directly draws the final cooked dessert, missing the intermediate coherent steps of generation. These coherent generation steps not only enhance interpretability but also boost the reasoning capabilities of the model~\cite{wei2022chain}.
2) \textbf{Entity and Style Consistency}: The entity and subject styles of the generated illustrated images are inconsistent. As illustrated in Figure~\ref{fig:intro_generation_results_comparison}(b), the style of the blue-gray sofa image generated by the model pre-trained on MMC4 is inconsistent with that in the query. This consistency of entities and style has been proposed as an important metric for evaluating the results of interleaved image-text content generation~\cite{an2023openleaf}. These limitations primarily stem from the low quality of training data caused by the poor alignment between image and text content, \ie, \emph{the devil is in the data}. For example in Figure~\ref{fig:intro_generation_results_comparison}(c), the second image in the training sample has a noticeable gap with the context. Additionally, the styles of all the images are almost completely different. Thus, it raises a natural research question: \textbf{\emph{How to construct a high-quality interleaved image-text dataset?}}

In this paper, we introduce \datasetname, a high-quality and coherent interleaved image-text dataset. By ``high-quality'', we mean that the interleaved image-text content should exhibit \emph{developed and logically structured text steps}, \emph{consistent entities and visual styles in images}, and \emph{strong semantic and contextual alignment between them}. 
% \todo{our filter strategies.} 
To enhance text coherence and image consistency, we collect raw data from specific websites, \eg, WikiHow~\cite{wikihow}, known for their instructional content or visual stories. As illustrated in Figure~\ref{fig:intro_generation_results_comparison}(d), instructional content, due to its uniform intention (\eg, cooking steak) and structured presentation (\eg, step 1), exhibits strong text coherence and image consistency. Additionally, we devise a multi-perspective filter strategy consisting of: a text sequence filter, an image sequence filter, and an image-text alignment filter. The first two leverage large language models (LLMs), \eg, Llama3~\cite{meta2024llama3}, and vision-language models (VLMs), \eg, CLIP~\cite{radford2021learning}, to eliminate incoherent text and images, respectively, further enhancing overall coherence. The last one employs both VLMs and MLLMs to comprehensively assess and refine the dataset for better alignment between visual and textual elements. After constructing this dataset, we further enhance it into a preference dataset by generating negative samples through shuffled text and images. This preference dataset can then be used during the reinforcement learning stage of post-training~\cite{kaufmann2023survey, gao2024towards}.

To evaluate the data quality and make a comprehensive comparison of existing interleaved image-text benchmarks~\cite{laurenccon2024obelics, zhu2024multimodal}, we design four metrics to evaluate the \textbf{development}, \textbf{completeness}, \textbf{image-text alignment}, and \textbf{consistency of image-text sequences}, separately (\cf~Table~\ref{table:comp_with_previous}). \datasetname~significantly outperforms previous datasets~\cite{laurenccon2024obelics, zhu2024multimodal} on all metrics, demonstrating its high quality. To further demonstrate the superiority of \datasetname, we conduct few-shot experiments on various downstream tasks (\eg, VQA)~\cite{goyal2017vqav2, gurari2018vizwiz, kiela2020hateful, lin2014coco, marino2019okvqa}, comparing with MMC4~\cite{zhu2024multimodal} and OBELICS~\cite{laurenccon2024obelics}. The model trained on our \datasetname~consistently outperforms others, particularly in long context settings (16 \& 32 shots), showcasing \datasetname's ability to enhance the in-context learning capacities of MLLMs.

Furthermore, thanks to our dataset, we introduce four new challenging tasks to evaluate the multimodal understanding and generation abilities of MLLMs: \textbf{image-to-text sequence generation}, \textbf{text-to-image sequence generation}, \textbf{interleaved image-text content continuation}, and \textbf{question-based interleaved image-text generation}. 
To assess the performance of these newly introduced tasks, we implement a comprehensive evaluation framework considering metrics such as METEOR~\cite{banerjee2005meteor}, ROUGE~\cite{lin2004rouge} for text quality, and FID~\cite{heusel2017fid} and IS~\cite{salimans2016improved} for image quality. To accurately evaluate open-ended interleaved image-text generation content, we develop a series of metrics and leverage powerful MLLM (\eg, GPT4o~\cite{openai2024gpt4o}) for assessment. These tasks and evaluations pave the way for novel approaches in evaluating MLLMs by addressing critical gaps in current benchmarks.

Conclusively, our contributions are as follows:

\vspace{-0.3em}
\begin{itemize}[leftmargin=*]
\itemsep-0.2em

    \item[1)] We introduce \datasetname, a high-quality coherent interleaved image-text dataset designed to address the limitations of existing datasets by ensuring coherent narrative, consistent entity and style, and strong semantic alignment between images and text.
    \item[2)] \datasetname~introduces four novel benchmark tasks and associated evaluation metrics to comprehensively evaluate multimodal understanding and generation capabilities of MLLMs. 
    \item[3)] \datasetname~can serve as both a supervised fine-tuning dataset and a preference dataset. Extensive ablations across various downstream tasks demonstrate the high-quality of \datasetname~and its effectiveness in enhancing the understanding and generation capabilities of MLLMs.
\end{itemize}

\section{Related Work}

\noindent\textbf{Interleaved Image-Text Web Document Datasets.}
Training on interleaved image-text web documents has demonstrated superior performance over image-description pairs, as evidenced by studies such as Flamingo~\cite{alayrac2022flamingo} and KOSMOS-1~\cite{huang2302language}. These findings underscore the significant advantages of utilizing the richer and more meaningful correlations inherent in interleaved image-text documents\cite{laurenccon2024obelics}. However, the training data in these two studies is not made publicly available.

The scarcity of interleaved image-text data has been addressed by the introduction of two datasets: MMC4~\cite{zhu2024multimodal} and OBELICS~\cite{laurenccon2024obelics}. MMC4 builds upon the text-only C4 corpus~\cite{raffel2020exploring} by integrating images into text passages using CLIP features.
% , resulting in a dataset comprising 101.2 million documents, 571 million images, and 43 billion English tokens. 
In contrast, OBELICS emphasizes comprehensive filtering strategies and preserves the original structure of web pages. 
% It encompasses 141 million web pages, 353 million images, and 115 billion text tokens.\todo{Add our data statics.}
These datasets present a diverse and extensive interleaved image-text corpus for multimodal language models. Despite their depth and breadth, they exhibit notable deficiencies in document quality, such as weaker completeness and image-text coherence (\cf, Table~\ref{table:comp_with_previous}). Additionally, as depicted in Figure~\ref{fig:data_distribution}, their documents lack comprehensiveness in the image modality, with most documents containing only one or two images. To address this, we propose CoMM that focuses explicitly on coherence-rich scenarios with more image illustrations. CoMM greatly enhances the baseline's ~\cite{awadalla2023openflamingo} performance (in Table~\ref{table:perf_flamingo}), highlighting the critical role of a high-quality coherent interleaved image-text training corpus.

\noindent\textbf{Modeling of Interleaved Image-Text Data.} 
The impressive performance of language modeling by LLM motivates researchers to delve deeper into the comprehension and generation capabilities of multimodal data.  Flamingo~\cite{alayrac2022flamingo} introduces image tokens into the language modeling process, facilitating interleaved image-text input and enabling few-shot transfer to tasks (\eg, VQA). 
Emu~\cite{sun2023emu} leverages Stable Diffusion~\cite{rombach2022ldm} as the image decoder, enabling the generation of image or text from interleaved image-text input. DreamLLM~\cite{dong2023dreamllm} pioneers the generation of free-form interleaved image-text outputs by modeling text and image within a unified multimodal space.

Recent research has increasingly focused on the intrinsic requirement of interleaved image-text modeling: capturing multimodal coherence while generating interleaved image-text content. OpenLEAF~\cite{an2023openleaf} underscores the importance of maintaining semantic and stylistic consistency between generated images. It also proposes a training-free baseline and employs BingChat~\cite{bingchat} to evaluate the two types of consistency. MM-Interleaved~\cite{tian2024mm} introduces the multimodal feature synchronizer to extract context-sensitive, multi-scale image features, thereby more effectively capturing multimodal context coherence and visual consistency. 
Nonetheless, focusing solely on the model level to meet the intrinsic multimodal coherence requirement is insufficient. Therefore, our high-quality data that strictly adhere to this requirement are essential to fully realize the potential of MLLMs.

\section{\datasetname~Dataset}

% \subsection{Overview}
% \lil{Overview}

The \datasetname~dataset is collected from high-quality interleaved image-text content sourced from various websites, focusing on instructional steps and visual stories. 
Besides, we apply multi-perspective quality filter strategies on text sequences, image sequences, and image-text alignment, leveraging advanced models, \eg, CLIP~\cite{radford2021learning} and LLMs~\cite{meta2024llama3}, to enhance the dataset's coherence and relevance\footnote{Due to space constraints, more details are left in the Appendix.\label{foot:appendix}}. 
% Further s on the collection and filtering processes are provided in the Appendix. 
% \todo{Dataset size, license, and checklist}

\subsection{Dataset Construction} \label{sec:dataset_construct}
% \subsection{Collection Process} 
\label{ref:collect_process}

\noindent\textbf{Collection Process.}
As mentioned above, we gather specific interleaved image-text data to preliminarily ensure the coherence and consistency of our dataset.
To achieve this, we explore websites that are likely to provide high-quality interleaved content, such as instructional steps and visual stories. We collect raw data from these sources\footref{foot:appendix} and download the corresponding text and images. Initially, we apply basic data cleaning techniques, including de-duplication and filtering out images with low resolution, to refine the collected data.

\textit{Discarding NSFW images.} To address potential ethical implications, we follow the~\cite{zhu2024multimodal} by utilizing an advanced NSFW binary image classifier~\cite{gadre2024datacomp}. This classifier is devised as a 4-layer multi-layer perceptron (MLP) that has been fully trained on the NSFW dataset from LAION-2B~\cite{schuhmann2022laion}. Specifically, it utilizes features extracted from OpenAI’s CLIP ViT-L/14 model\cite{radford2021learning} as input, achieving an impressive accuracy of 97.4\% on the NSFW test dataset. For each image processed, we employ the classifier to predict the NSFW probability. Images yielding a probability greater than 0.1 are automatically excluded from further consideration.

% \subsection{Qualities Filter Strategy}
\noindent\textbf{Qualities Filter Strategy.} Due to the inherent noises in the gathered raw data, the data still contains inconsistent and irrelevant content. Thus, we further employ a multi-perspective filter strategy to ensure the quality and coherence of the dataset. Specifically, it involves three following components\footref{foot:appendix}:

\textit{1) Text Sequence Filter. } 
For text sequences, the most important aspect is the contextual development and cohesive connection between them. To maintain a smooth and logical flow of the text, we employ LLMs (\eg, Llama3~\cite{meta2024llama3}) to evaluate the \textbf{development} and \textbf{coherence} among text steps in a document (prompt details are in the Appendix). According to evaluation scores, we eliminate sequences that lack coherence and relevance in their content, producing a higher-quality text corpus.

\textit{2) Image Sequence Filter.} 
As for image sequences, visual consistency and relevance are key factors in maintaining the quality of the content. To assess the coherence and development of image sequences, we devise a metric $\mathcal{F}(\cdot)$ through the CLIP's visual encoder~\cite{radford2021learning} in Eq.~(\ref{eq:image_seq}):
\begin{align}
    \mathcal{F} (\{x_{i} | 1 \leq i \leq N\}) = \frac{1}{N - 1} \sum_{i=2}^{N} \text{Sim} (x_{i}, x_{i-1}) & \notag \\
    - \frac{2}{(N-1)(N-2)} \sum_{i=2}^{N} \sum_{j=1}^{i-1} \text{Sim} (x_{i}, x_{j}),
    \label{eq:image_seq}
\end{align}
where $\{x_{i} | 1 \leq i \leq N \}$ denotes the sequence of images from 1 to $N$, $\text{Sim}(\cdot, \cdot)$ denotes the cosine similarity score between two images calculated by the CLIP's visual encoder. The first term denotes the average similarity between consecutive images, promoting smooth transitions and visual coherence within the sequence, while the second term represents the average similarity between all pairs of images, ensuring overall diversity and development. In this way, we can effectively balance coherence and diversity, ensuring that the image sequence is both visually consistent and contextually developed.
% \lil{The first item denotes xxx, while the second xxx. By this way, we can xxxxxx}
By applying the image sequence filter, we can identify and eliminate sequences that lack connection or development, resulting in more polished and visually appealing images.

\textit{3) Image-Text Alignment Filter.} This filter aims to ensure that the images and text are not only relevant to each other but also maintain a coherent narrative throughout the document. 
We first utilize CLIP~\cite{radford2021learning} to calculate the CLIP similarity scores between images and text and remove those whose scores are less than 0.1. Since CLIP is trained on individual image-text pairs, where the text often serves as the image's caption, it tends to favor concrete descriptions (\ie, object descriptions and attributes)~\cite{lavoie2024modeling}. However, in an interleaved image-text document, the text for each step describes progress with associated images (\eg,  in Figure~\ref{fig:intro_generation_results_comparison}(d), each text focuses on a cooking action linked to previous steps) which is associated with previous steps. 
Therefore, we also employ advanced model (\eg, GPT-4o~\cite{openai2024gpt4o} or Llama3~\cite{meta2024llama3}) to assess and filter image-text pairs, as these models can better judge current image-text alignment by considering the previous context.

\noindent\textbf{Preference Dataset Construction.} Preference learning~\cite{gao2024towards} has recently gained popularity in the post-training of LLMs, leveraging reinforcement learning~\cite{kaufmann2023survey} to train models with both positive and negative samples. In our dataset, which consists of interleaved image and text sequences, negative samples are easily created by shuffling the order of steps. These shuffles can be categorized into four types:
\textit{Shuffled Text}: Text in different steps is shuffled, while images remain in their original order.
\textit{Shuffled Image}: Images are shuffled, while the text remains in its original order.
\textit{Both Shuffled Text and Image}: Both text and images are shuffled independently.
\textit{Shuffled Steps}: Images and text within each step remain aligned, but the sequence of steps is randomized.

These negative samples are combined with original positive samples to form a preference dataset, used for further model training following supervised fine-tuning.

\begin{figure*}[!htbp]
    \centering
    \centering
    \includegraphics[width=\textwidth]{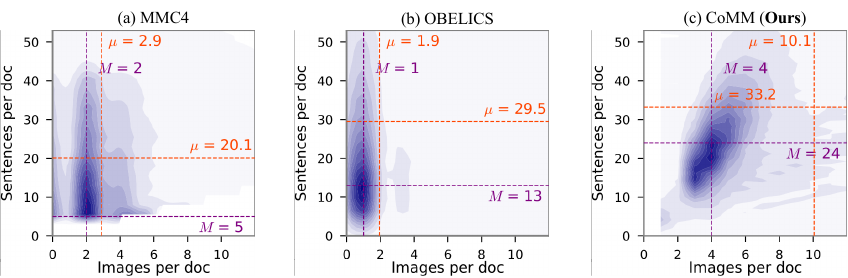}
    \vspace{-2em}
    \caption{
    Visualization of the image-sentence numbers per document distribution of three datasets. The $\mu$ and $M$ denote the mean/median number of images/sentences in documents, respectively.}
    \label{fig:data_distribution}
    \vspace{-1em}
\end{figure*}

\subsection{Dataset Statistic}
After filtering, our dataset contains 227K documents with 2.28M images. 
% \todo{We provide more details in the Appendix (more visualization, source, \etc)}
Further, Figure~\ref{fig:data_distribution} illustrates the distribution of image-sentence number per document across MMC4, OBELICS, and CoMM datasets. Darker colors indicate a higher number of documents. The $\mu$ and $M$ represent the average and the most common number of images/sentences per document, respectively. For MMC4 and OBELICS, the distribution of images per document is highly concentrated around very low values of $M$ (2 and 1 per document, respectively). This indicates a significant scarcity of the image modality in their multimodal data, which directly limits the models' multimodal in-context learning ability.
% which directly limits the models' ability to leverage multimodal context for learning to understand and generate images.

Examining Figure~\ref{fig:data_distribution}(a) and Figure~\ref{fig:data_distribution}(b) distributions, it is evident that the number of images does not significantly change with an increase in the number of sentences. However, in practical scenarios, longer documents are more likely to contain a greater number of illustrations, which aids in modeling longer multimodal contexts. In contrast, CoMM features a larger quantity of images per document ($M$ = 4, $\mu$ = 10.1), which significantly enhances the dataset's image modality while focusing on multimodal coherence. Furthermore, CoMM exhibits a trend where longer documents tend to include more image illustrations, thereby facilitating the modeling of longer multimodal contexts.

\section{Interleaved Generation Task and Benchmark}

\subsection{Task Formulation}
In this section, we propose four benchmark tasks to evaluate MLLMs' interleaved generation capabilities. For ease of presentation, we utilize 
$\{x_{i} | 1 \leq i \leq N\}$ and $\{t_{i} | 1 \leq i \leq N\}$ to represent the image sequences and text sequences, respectively. Besides, we use $t_{\text{prompt}}$ as the text prompt, and $\mathcal{F}_{\theta} (\cdot)$ as the model with parameter $\theta$. The detailed formulations of tasks are outlined as follows:

\textbf{Task 1: Image-to-Text Sequence Generation.}
This task evaluates the model's capability to generate coherent and contextually appropriate narratives based on sequential images. The objective is to predict the corresponding text sequences from the given image sequences:
\begin{equation} 
\small
    \{t_{i} | 1 \leq i \leq N\} = \mathcal{F}_{\theta} (t_{\text{prompt}}, x_{1}, ... , x_{N}).
\end{equation}
\textbf{Task 2: Text-to-Image Sequence Generation.}
This task assesses the model's proficiency in generating image sequences that accurately represent provided textual descriptions. The goal is to predict the corresponding image sequences from the given text sequences:
\begin{equation} 
\small
    \{x_{i} | 1 \leq i \leq N\} = \mathcal{F}_{\theta} (t_{\text{prompt}}, t_{1}, ... , t_{N}).
\end{equation}
\textbf{Task 3: Interleaved Image-Text Content Continuation.}
It measures the model's ability to seamlessly continue content in an interleaved image-text format. The model needs to predict the outputs of the remaining image-text steps based on the interleaved image-text content in the previous $k$ steps:
\begin{equation} 
\small
    \{t_{i}, x_{i} | k < i \leq N\} = \mathcal{F}_{\theta} (t_{\text{prompt}}, t_1, x_1, ... , t_{k}, x_{k}).
\end{equation}
\textbf{Task 4: Question-based Interleaved Image-Text Generation.}
\label{sec:task4}
This challenging task requires the model to generate interleaved content based on a given question, testing its understanding and generative reasoning abilities across modalities. The model needs to predict the outcomes of all image-text steps based on the given question:
\begin{equation} 
\small
    \{t_{i}, x_{i} | 1 \leq i \leq N\} = \mathcal{F}_{\theta} (t_{\text{prompt}}).
\end{equation}
\subsection{Evaluation Metric}
This section lists the evaluation metrics for assessing the quality of generated results. Our evaluation framework\footref{foot:appendix} measures textual accuracy, visual fidelity, and the coherence of multi-modal outputs.

\textbf{Evaluation for Text Sequence Generation.}
Generating narratives for a series of images shares similarities with the established task of image captioning. Consequently, we directly apply traditional image captioning metrics like \textit{ROUGE}~\cite{lin2004rouge} and \textit{METEOR}~\cite{banerjee2005meteor} at both the step and document levels. 
Additionally, we propose \textit{Illustration Relevance Score} (IRS) to assess the relevance between image illustration and text context in the final document. Specifically, we simulate the human document creation process by first generating detailed image descriptions from the text context and then assessing the consistency between these descriptions and corresponding images using GPT-4o~\cite{openai2024gpt4o}.

\textbf{Evaluation for Image Sequence Generation. }
We evaluate the quality of image illustrations on four key assessment aspects including image quality, style consistency between generated images, reconstruction on referenced images, and \textit{IRS}.
%, and the coherence of the interleaved image-text sequence. 
To evaluate image quality, we use \textit{Fréchet Inception Distance} (FID)~\cite{heusel2017fid} with ground truth reference and \textit{Inception Score} (IS)~\cite{salimans2016improved}. For assessing style consistency, we employ the \textit{Structural Similarity Index Measure} (SSIM), and for ground truth reconstruction evaluation, we adopt the \textit{Peak Signal-to-Noise Ratio} (PSNR).

\textbf{Evaluation for Interleaved Image-Text Generation.}
We use MLLM (GPT-4o~\cite{openai2024gpt4o}) for a comprehensive assessment of linguistic and visual coherence. GPT-4o evaluates four aspects: image sequence coherence, image quality, document completeness, and \textit{IRS}.
For image sequence coherence, we check \textit{style}, \textit{entity} consistency, and content \textit{trend} alignment between image and text sequence. \textit{Image quality} is evaluated on authenticity, integrity, clarity, and aesthetics. Document \textit{completeness} is scored based on how thoroughly the content covers the topic.

\section{Experiments}

\subsection{Datasets Quality Comparison}
\textbf{Settings.} In this section, we compare the quality of our dataset with previous interleaved image-text datasets (\eg, MMC4~\cite{zhu2024multimodal} and OBELICS~\cite{laurenccon2024obelics}) focusing on text sequences, image sequences, and image-text alignment. We utilize powerful LLMs (\eg, Llama3~\cite{meta2024llama3} and GPT-4o~\cite{openai2024gpt4o}) to assess the text development, text completeness, and image-text alignment of these datasets. Each metric is scored on a scale of 0 to 10, with detailed definitions and prompts provided in the Appendix. For MMC4 and OBELICS, we randomly sample 5,000 documents from their data. We evaluate our entire dataset with Llama3 and randomly sample 5,000 documents for assessment by GPT-4o. To evaluate image sequence quality, we employ the metric $\mathcal{F}$ as cited in Eq.~\eqref{eq:image_seq}.

\textbf{Quantitative Results.} As shown in Table~\ref{table:comp_with_previous}, our dataset surpasses MMC4 and OBELICS across all evaluated dimensions. Notably, the score difference in development, completeness, and image-text alignment is more pronounced when evaluated by GPT-4o, a more powerful MLLM. Furthermore, the image sequence metric underscores the superior quality of our data, reflecting higher scores compared to MMC4 and OBELICS. These results affirm the high quality of our dataset in interleaved image-text data.

\begin{table}[t]
% \begin{wraptable}[10]{r}{7.5cm}
\tabcolsep=1.5mm
\centering
% \vspace{-1.3em}

% \vspace{-1mm}
\small
\begin{tabular}{c|cccc}
% \noalign{\hrule height 1.5pt}
\specialrule{0.05em}{0pt}{0pt}
\hline
 \multirow{2}{*}{Datasets} & DLP &  CPL & ITA & \multirow{2}{*}{ImgS} \\
          & L.3 / G.4o &     L.3 / G.4o      &     L.3 / G.4o      &         \\
\hline
 MMC4~\cite{zhu2024multimodal} & 5.56 / 4.75 & 6.28 / 5.12 & 6.53 / 4.66 &  0.21  \\
 OBELICS~\cite{laurenccon2024obelics} & 5.60 / 5.97 & 6.51 / 5.88 & 5.00 / 3.81 & 1.00  \\
 \datasetname\ (\textbf{Ours}) & \textbf{6.93} / \textbf{7.64} & \textbf{7.44} / \textbf{7.07} & \textbf{7.46} / \textbf{8.91} & \textbf{4.27}  \\

\specialrule{0.05em}{0pt}{0pt}
\hline
\end{tabular}
\caption{Quality comparison of interleaved image-text datasets. ``DLP'' stands for Development, ``CPL'' signifies Completeness, ``ITA'' represents Image-Text Alignment, ``ImgS'' refers to Image Sequence, ``L.3'' denotes Llama3~\cite{meta2024llama3}, and ``G.4o'' is GPT-4o~\cite{openai2024gpt4o}. }
\label{table:comp_with_previous}
\end{table}
% \end{wraptable}

\subsection{In-Context Multimodal Understanding}

% \textcolor{red}{experimetnal settings more clear.}

\noindent\textbf{Setting. }
To further demonstrate the high-quality and validate the efficacy of our dataset in promoting in-context understanding, we conduct experiments using few-shot learning settings on several downstream tasks (COCO~\cite{lin2014coco}, Flickr30k~\cite{young2014flickr30k}, VQAv2~\cite{goyal2017vqav2}, OKVQA~\cite{marino2019okvqa}, TextVQA~\cite{singh2019textvqa}, VizWiz~\cite{gurari2018vizwiz}, and HatefulMemes~\cite{kiela2020hateful}). Few-shot learning is critical for assessing a model's ability of in-context understanding.
In our experiments, few-shot scenarios are simulated by providing the model with a small number of example queries and their corresponding answers before it is evaluated on new queries. We consider various shot settings (0, 4, 8, 16, and 32) to understand the model's performance across different levels of supervision.
% ~\cite{goyal2017vqav2, gurari2018vizwiz, kiela2020hateful, lin2014coco, marino2019okvqa, singh2019textvqa, young2014flickr30k}

\noindent\textbf{Quantitative Results.}
Table \ref{table:perf_flamingo} presents the results for the baseline model and its enhancements with MMC4, OBELICS, and our proposed dataset. As observed from Table~\ref{table:perf_flamingo}, our method consistently outperforms the performance of MMC4 and OBELICS across all shot settings, demonstrating superior generalization and in-context learning capabilities.
Particularly noticeable improvements are seen in the COCO and TextVQA datasets, where our method exceeds the baseline by significant margins in nearly all shot settings. This superior performance underlines the strength of our dataset in enabling models to better learn from context and apply learned knowledge to unseen instances. Furthermore, our dataset surpasses the baseline in all tasks with more shot settings, maintaining its advantage with longer context few-shot data, as shown in the 16-shot and 32-shot scenarios.

% \addtolength{\tabcolsep}{-1pt}
\begin{table*}[]
\centering

% \vspace{0.5em}
\small
\begin{tabular}{lc|ccccccc}
\specialrule{0.05em}{0pt}{0pt}
\hline
% & \rotatebox[origin=c]{45}{Shot} & \rotatebox[origin=c]{45}{COCO} & \rotatebox[origin=c]{45}{Flickr30k} & \rotatebox[origin=c]{45}{VQAv2} & \rotatebox[origin=c]{45}{OKVQA} & \rotatebox[origin=c]{45}{TextVQA} & \rotatebox[origin=c]{45}{VizWiz} & \rotatebox[origin=c]{45}{HatefulMemes} \\ 
& Shot & COCO & Flickr30k & VQAv2 & OKVQA & TextVQA & VizWiz & HatefulMemes \\ 
\hline
% Flamingo-9B & \multirow{5}{*}{0} & - & 79.4 & 61.5 & 51.8 & \textbf{44.7} & 31.8 & 22.8 &  57.0 \\
Baseline & \multirow{4}{*}{0} &  79.5 & 59.5 & \textbf{52.7} & 37.8 & 24.2 & \textbf{27.5} &  51.6 \\
% IDEFICS-9B &  & - & 46.0 & 27.3 & 50.9 & 38.4 & 25.9 & \textbf{35.5} & 51.8 \\ 
 + MMC4~\cite{zhu2024multimodal} &    & 97.1 & 67.0 & 45.8 & 29.8 & 23.4 & 15.0 & 54.6 \\
% + MMC4(filtered) &  & 16 &  &  &  &  &  &  &  \\
+ OBELICS~\cite{laurenccon2024obelics} &    & 83.5 & 51.5 & 50.9 & 38.9 & 25.6 & 21.9 & 53.8 \\
+ CoMM (\textbf{Ours}) &   & \textbf{100.3} & \textbf{69.8} & 52.6 & \textbf{41.3} & \textbf{33.5} & 23.6 & \textbf{54.8} \\
\hline
% Flamingo-9B & \multirow{5}{*}{4} & - & 93.1 & 72.6 & \textbf{56.3} & \textbf{49.3} & {33.6} & \textbf{34.9} & \textbf{62.7} \\
Baseline & \multirow{4}{*}{4} & 89.0 & 65.8 & \textbf{54.8} & 40.1 & 28.2 & \textbf{34.1} & 54.0 \\
% IDEFICS-9B &  & - & 93.0 & 59.7 & 55.4 & 45.4 & 27.6 & 36.9 & 50.7 \\ 
+ MMC4~\cite{zhu2024multimodal} &  &  103.3 & 68.8 & 49.5 & 36.7 & 28.7 & 24.4 & 55.0 \\
% + MMC4(filtered) &  & 16 &  &  &  &  &  &  &  \\
+ OBELICS~\cite{laurenccon2024obelics} &  &  103.7 & 65.2 & 53.2 & 43.0 & 31.4 & 29.9 & 55.7 \\
 + CoMM (\textbf{Ours}) &  &  \textbf{107.0} & \textbf{73.2} & 53.6 & \textbf{42.9} & \textbf{35.7} & 30.1 & \textbf{55.2} \\
\hline
% Flamingo-9B & \multirow{5}{*}{8} & - &99.0 & {73.4} & \textbf{58.0} & \textbf{50.0} & {33.6} & 39.4 & 63.9 \\
Baseline & \multirow{4}{*}{8} & 96.3 & 62.9 & \textbf{54.8} & 41.1 & 29.1 & \textbf{38.5} & \textbf{54.7} \\
% IDEFICS-9B &  & - & 97.0 & 61.9 & 56.4 & 47.7 & 27.5 & 40.4 & 51.1 \\ 
+ MMC4~\cite{zhu2024multimodal} &   & 103.6 & 67.9 & 50.5 & 39.0 & 28.8 & 32.7 & 51.3 \\
% + MMC4(filtered) &  & 16 &  &  &  &  &  &  &  \\
+ OBELICS~\cite{laurenccon2024obelics} &   & 101.9 & 63.9 & 53.7 & \textbf{44.5} & 32.2 & 37.2 & 51.2 \\
+ CoMM (\textbf{Ours}) &   & \textbf{108.0} & \textbf{74.8} & 54.2 & \textbf{44.5} & \textbf{35.4} & 37.9 & 52.5 \\
\hline
% Flamingo-9B & \multirow{5}{*}{16} & - & 102.2 & 72.7 & 59.4 & 50.8 & 33.5 & 43.0 & {64.5} \\
Baseline & \multirow{4}{*}{16} & 98.8 & 62.8 & \textbf{54.3} & 42.7 & 27.3 & 42.5 & 53.9 \\
% IDEFICS-9B &  & - & 99.7 & 64.5 & 57.0 & 48.4 & 27.9 & 42.6 & 50.1 \\ 
+ MMC4~\cite{zhu2024multimodal} &   & 102.9 & 65.5 & 50.0 & 39.1 & 26.4 & 36.9 & 57.2 \\
% + MMC4(filtered) &  & 16 &  &  &  &  &  &  &  \\
+ OBELICS~\cite{laurenccon2024obelics} &   & 100.8 & 58.0 & 53.1 & 45.3 & 30.2 & 42.4 & 57.8 \\
+ CoMM (\textbf{Ours}) &   & \textbf{109.3} & \textbf{71.5} & {54.2} & \textbf{46.9} & \textbf{35.8} & \textbf{43.5} & \textbf{59.9} \\
\hline
% Flamingo-9B & \multirow{5}{*}{32} & - & {106.3} & 72.8 & {60.4} & {51.0} & 32.6 & {44.0} & 63.5 \\
Baseline & \multirow{4}{*}{32} & 99.5 & 61.3 & 53.3 & 42.4 & 23.8 & {44.0} & 53.8 \\
% IDEFICS-9B &  & - & 98.0 & 64.3 & 57.9 & 49.6 & 28.3 & 43.7 & 49.8 \\
+ MMC4~\cite{zhu2024multimodal} &   & 94.9 & 56.4 & 48.3 & 35.6 & 20.0 & 37.7 & 49.2 \\
% + MMC4(filtered) &  & 16 &  &  &  &  &  &  &  \\
+ OBELICS~\cite{laurenccon2024obelics} &   & 96.5 & 53.1 & 52.0 & 43.6 & 26.8 & 44.5 & 56.1 \\
+ CoMM (\textbf{Ours}) &   & \textbf{111.9} & \textbf{70.7} & \textbf{54.0} & \textbf{46.6} & \textbf{34.9} & \textbf{45.7} & \textbf{57.3} \\
\specialrule{0.05em}{0pt}{0pt}
\hline
\end{tabular}
\caption{Performance comparison of different datasets, with the baseline model Open-Flamingo 9B~\cite{awadalla2023openflamingo}. Evaluations were in an open-ended setting for VQA tasks with random in-context examples. (Task, Metric, Query Split): (COCO~\cite{lin2014coco}, CIDEr, test), (Flickr30k~\cite{young2014flickr30k}, CIDEr, test (Karpathy)), (VQAv2~\cite{goyal2017vqav2}, VQA accuracy, testdev), (OKVQA~\cite{marino2019okvqa}, VQA accuracy, val), (TextVQA~\cite{singh2019textvqa}, VQA accuracy, val), (VizWiz~\cite{gurari2018vizwiz}, VQA accuracy, testdev), (HatefulMemes~\cite{kiela2020hateful}, ROC-AUC, test seen). }
\label{table:perf_flamingo}
\end{table*}
\addtolength{\tabcolsep}{1pt}

% \input{tables/task1_results}
% \input{tables/task2_results}
% \input{tables/task3_results}

% \addtolength{\tabcolsep}{-3pt}
\begin{table*}[t]
% \tabcolsep=1.5mm
% \vspace{-1em}
\centering

% \vspace{0.5em}
\small
\begin{tabular}{lc|ccccc|cccccc}
% \noalign{\hrule height 1.5pt}
\specialrule{0.05em}{0pt}{0pt}
\hline
 \multirow{2}{*}{Methods} & \multirow{2}{*}{Size} & \multicolumn{5}{c|}{I2T Sequence Generation (Task1)} &  \multicolumn{6}{c}{Continuation Generation (Task3)} \\
        &  & M.S. &      R.S.     &     M.W.    &     R.W. & IRS &  Style  &   Entity  &    Trend & CPL. & ImgQ & IRS \\
\hline
 MiniGPT-5~\cite{zheng2023minigpt} & 7B & \textbf{4.60} & \textbf{8.64} & \textbf{9.10}  & \textbf{10.67} &  7.39 & 5.58 & 5.21 & 5.24  & 6.33  & 6.36 & 2.56 \\
SEED-Llama~\cite{ge2023making} & 8B &  2.29 & 6.34 & 4.20  & 7.33 & 6.99 & 6.28 & 5.84 & 5.72  & 6.28 & 6.55 & 2.92  \\
SEED-Llama~\cite{ge2023making} & 14B &  3.16 & 8.12 & 5.71  & 8.17 &  \textbf{8.13} & 6.68 & 6.22 & 6.13 & 6.66  & 6.67 & \textbf{3.23} \\
 Emu2~\cite{sun2023generative}& 33B & 3.65 & 8.41 & 7.43 & 8.62 &  5.01 & \textbf{8.22} & \textbf{7.99} & \textbf{7.97}  & \textbf{8.49} & \textbf{8.62} & 2.42 \\
\hline
 \multirow{2}{*}{Methods} & \multirow{2}{*}{Size} & \multicolumn{5}{c|}{T2I Sequence Generation (Task2)} & \multicolumn{6}{c}{ Question-based  Generation (Task4)} \\
        &  & FID $\downarrow$ &  IS &  SSIM &  PSNR  & IRS  & Style  &   Entity  &    Trend & CPL. & ImgQ & IRS \\
\hline
 MiniGPT-5~\cite{zheng2023minigpt} & 7B & \textbf{56.51} & 7.60 & 20.66 & 7.57 & 5.48 &  5.65 & 5.20 & 5.25 & 5.81 & 6.15 & \textbf{2.71} \\
 SEED-Llama~\cite{ge2023making} & 8B & 57.96 & 8.40 & 20.53 & 7.87  & 5.27  & 7.55 & 6.81 & 6.15  & 5.13 &6.36 & 1.46 \\
 SEED-Llama~\cite{ge2023making} & 14B &  66.23 & \textbf{11.06} & \textbf{20.83} & \textbf{8.12} & \textbf{6.24} & 7.51  & 6.61 & 6.30 & 6.13 & 6.66  & 2.50 \\
 Emu2~\cite{sun2023generative}& 33B & 60.45 & 6.34 & 20.96 & 7.91 & 4.17  & \textbf{8.41}  & \textbf{7.56} & \textbf{7.63} & \textbf{7.54} & \textbf{7.59} & 2.02 \\
\specialrule{0.05em}{0pt}{0pt}
\hline
\end{tabular}
\caption{Comparison of performance among MiniGPT-5~\cite{zheng2023minigpt}, SEED-Llama~\cite{ge2023making}, and Emu2~\cite{sun2023emu} on the four generation tasks. ``M.S.'' and ``R.S.'' represent METEOR~\cite{banerjee2005meteor} and ROUGE\_L \cite{lin2004rouge} for step-by-step evaluation, while ``M.W.'' and ``R.W.'' correspond to METEOR and ROUGE\_L for whole document evaluation. ``CPL.'' stands for Completeness, ``ImgQ'' indicates Image Quality, ``IRS'' means Illustration Relevance Score, and ``$\downarrow$'' denotes that lower values are better.}
\label{table:task1_reuslts}
% \vspace{-2mm}
\end{table*}
\addtolength{\tabcolsep}{3pt}

\subsection{Interleaved Generation Tasks}

\noindent\textbf{Baselines.}
This section introduces baselines developed from several state-of-the-art approaches, \ie, \textbf{MiniGPT-5}~\cite{zheng2023minigpt}, \textbf{SEED-Llama}~\cite{ge2023making}, and \textbf{Emu2}~\cite{sun2023generative}, to assess their effectiveness in generating interleaved image-text content. Details of these models are thoroughly provided in the Appendix. To construct the baselines, we fine-tune each of these models using our CoMM dataset. We employ a rigorous evaluation, assessing them on benchmark tasks involving interleaved image-text generation and understanding.

\begin{figure*}[!t]
    \centering
    \includegraphics[width=0.99\linewidth]{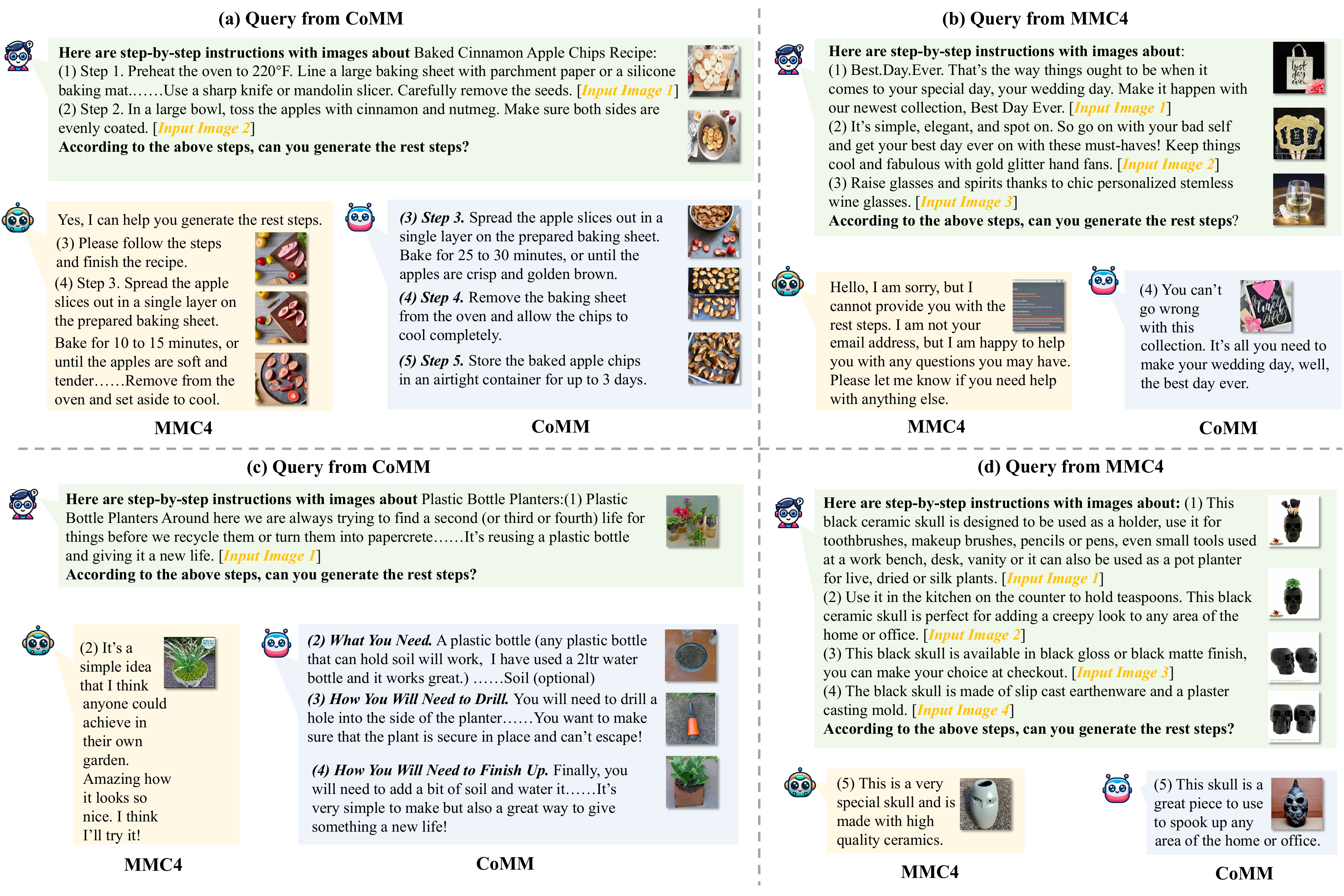}
    \vspace{-1.5em}
    \caption{Visualization of interleaved image-text content generation from SEED-Llama~\cite{ge2023making}~(Top) and MiniGPT-5~\cite{zheng2023minigpt}~(Bottom) separately trained by MMC4~\includegraphics[scale=0.2]{figures/mmc4_icon.png}~\cite{zhu2024multimodal} and \datasetname~\includegraphics[scale=0.2]{figures/ours_icon.png} (\textbf{Ours}). }
    \label{fig:comparison_visualization}
    \vspace{-1em}
\end{figure*}

\noindent\textbf{Quantative Results.}
As shown in Table~\ref{table:task1_reuslts}, the quantitative results clearly illustrate the varying performance of MiniGPT-5, SEED-Llama, and Emu2 across the four interleaved generation tasks. For I2T Sequence Generation (Task1), MiniGPT-5 achieves the highest scores in METEOR, ROUGE\_L,  but falls short in Illustration Relevance Score (IRS) where SEED-Llama (14B) leads with a score of 8.13. 
For T2I Sequence Generation (Task2), MiniGPT-5 excels with the lowest FID score of 56.51, while SEED-Llama (14B) surpasses IS, SSIM, PSNR, and IRS metrics, suggesting stronger visual fidelity and semantic similarity. 
Emu2 tops in Continuation Generation (Task3) with the highest scores in Style, Entity, Trend, Completeness, and Image Quality.
In Question-based Generation (Task4), Emu2 dominates in Style, Entity, Trend, Completeness, and Image Quality, while MiniGPT-5 manages to secure the lead in IRS.
Overall, these results highlight the strengths and weaknesses of each model in handling different aspects of interleaved image-text generation tasks, providing valuable insights for future research and model improvements.

\noindent\textbf{Preference Dataset Training Results.} With SEED-Llama~\cite{ge2023making} unifying image to discrete token prediction modeling, we can directly apply the mainstream preference learning algorithm DPO~\cite{rafailov2024direct} to train SEED-Llama on our preference dataset following supervised fine-tuning on our dataset. As shown in Table~\ref{table:dpo_reuslts}, performance further improves on the image-to-text sequence generation task after training with our preference dataset. Additional results on preference dataset training are available in the Appendix.

\noindent\textbf{Multimodal Generation Qualitative Analysis.} 
% As shown in Figure~\ref{fig:intro_generation_results_comparison} and Figure~\ref{fig:comparison_visualization}, compared with the interleaved image-text content generated by models~\cite{sun2023generative, zheng2023minigpt, ge2023making} trained on MMC4~\cite{zhu2024multimodal}, the content generated by models trained on our datasets are more consistent with previous input context. 
As illustrated in Figure~\ref{fig:intro_generation_results_comparison} and Figure~\ref{fig:comparison_visualization}, the interleaved image-text content generated by models~\cite{sun2023generative, zheng2023minigpt, ge2023making} trained on MMC4~\cite{zhu2024multimodal} shows a noticeable difference when compared to the content generated by models trained on our datasets. Our models produce content that is more consistent with the previous input context (\eg, textual coherence and entity consistency). By ensuring higher coherence, consistency, and alignment, CoMM sets a new standard for training datasets in the realm of multimodal large language models. 
More visualizations are available in the Appendix.

\begin{table}[t]
\tabcolsep=1.5mm
\centering

\small
\begin{tabular}{lc|ccccc}
% \noalign{\hrule height 1.5pt}
\specialrule{0.05em}{0pt}{0pt}
\hline
 \multirow{2}{*}{Methods} & \multirow{2}{*}{Size} & \multicolumn{5}{c}{I2T Sequence Generation (Task1)}  \\
        &  & M.S. &      R.S.     &     M.W.    &     R.W. & IRS \\
\hline
  SEED-Llama~\cite{ge2023making} & 8B &  2.29 & 6.34 & 4.20  & 7.33 & 6.99  \\
 +~~~~\textbf{DPO}~\cite{rafailov2024direct} & 8B & \textbf{4.09} & \textbf{8.83} & \textbf{6.95} & \textbf{10.24} &  \textbf{7.26}  \\
\hdashline  
SEED-Llama~\cite{ge2023making} & 14B &  3.16 & 8.12 & 5.71  & 8.17 &  8.13  \\
 +~~~~\textbf{DPO}~\cite{rafailov2024direct} & 14B & \textbf{4.62} & \textbf{10.15} & \textbf{8.08} & \textbf{10.30} &  \textbf{8.18}  \\
\specialrule{0.05em}{0pt}{0pt}
\hline
\end{tabular}
\vspace{-2mm}
\caption{Performance results of SEED-Llama~\cite{ge2023making} trained by DPO~\cite{rafailov2024direct} in our preference dataset. ``M.S.'' and ``R.S.'' represent METEOR~\cite{banerjee2005meteor} and ROUGE\_L \cite{lin2004rouge} for step-by-step evaluation, while ``M.W.'' and ``R.W.'' correspond to METEOR and ROUGE\_L for whole document evaluation. }
\label{table:dpo_reuslts}
\vspace{-4mm}
\end{table}
% \addtolength{\tabcolsep}{3pt}

\section{Conclusion}

In this paper, we present \datasetname, a high-quality interleaved image-text dataset designed to address the limitations of existing multimodal large language models (MLLMs) by ensuring superior text-image alignment and entity consistency. Through diverse interleaved image-text content, \datasetname\ significantly enhances MLLMs' in-context learning capabilities, as demonstrated by extensive few-shot experiments. Additionally, we introduce four novel tasks and a comprehensive evaluation framework to assess interleaved image-text generation abilities, setting new benchmarks for advancing multimodal understanding and generation in MLLMs.

\section*{Acknowledgment}
This work was supported by the Hong Kong SAR RGC Early Career Scheme (26208924), the National Natural Science Foundation of China Young Scholar Fund (62402408), and the HKUST Sports Science and Technology Research Grant (SSTRG24EG04), ACCESS – AI Chip Center for Emerging Smart Systems, sponsored by InnoHK funding, Hong Kong SAR.

{
    \small
    \bibliographystyle{ieeenat_fullname}
    \bibliography{main}
}

\clearpage
\setcounter{page}{1}
\maketitlesupplementary
\setcounter{section}{0}
\renewcommand\thesection{\Alph{section}}

\section*{Appendix}

This supplementary document is organized as follows:
\begin{itemize}
    \item Details of our dataset are shown in Sec.~\ref{sec:dataset_detail}.
    \item More experiment studies and analyses are shown in Sec.~\ref{sec:more_abla}.
    \item Ethical discussion and license are displayed in Sec.~\ref{sec:ethic_impact}.
    \item Filter strategy and evaluation details are in Sec.~\ref{sec:prompt_details}.
    \item Model training details are shown in Sec.~\ref{sec:model_training}. 
    \item More visualization results are illustrated in Sec.~\ref{sec:more_vis}.

\end{itemize}

\section{Details of Our Dataset}\label{sec:dataset_detail}
\textbf{Data Source. } As shown in Table~\ref{table:data_source}, we collect our data from five sources. For StoryGen~\cite{liu2023intelligent}, we use only the original source images and employ Llama3~\cite{meta2024llama3} to generate a more coherent and developed story text corpus. After filtering, CoMM contains 227K documents with 2.28M images and 139M text tokens. The ratio of different data sources is presented in Table~\ref{table:data_source}. We randomly sample 500 documents each for the validation and test splits, with the remaining documents used for training. 

\begin{table*}[h]
\centering
% \vspace{-1.3em}
\begin{tabular}{c|ccc}
\specialrule{0.05em}{0pt}{0pt}
\hline
 Data Source & Document Ratio & Image Ratio  & Text Token Ratio     \\
 \hline
 www.wikihow.com & 57.78\% & 30.02\% & 48.71\% \\
 www.ehow.com & 3.31\% & 4.11\% &  2.53\%  \\
 storybird.com & 1.54\% & 1.22\% &  0.57\%  \\
 StoryGen~\cite{liu2023intelligent} & 2.19\% & 1.76\% &  1.90\%  \\
 www.instructables.com & 35.18\% & 62.89\% &  46.28\%  \\
\hline

\specialrule{0.05em}{0pt}{0pt}
\hline
\end{tabular}
\caption{Collected data source of CoMM. }
\label{table:data_source}
\end{table*}

\begin{figure*}
    \centering
    \includegraphics[width=0.8\linewidth]{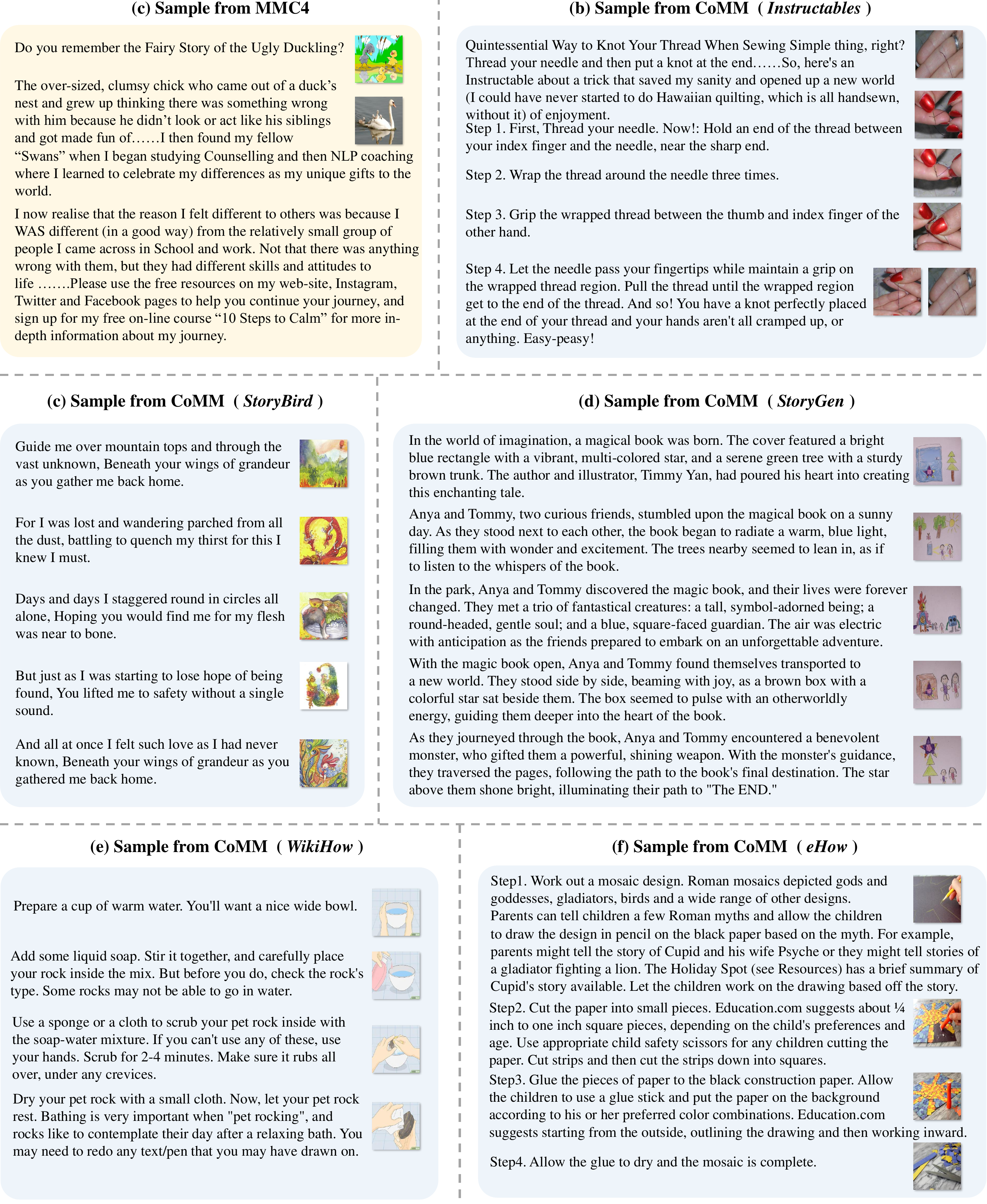}
    \caption{Comparison of samples from different datasets. (a) from the MMC4~\cite{zhu2024multimodal} dataset; (b)-(f) from different data sources within CoMM (\textbf{Ours}) dataset.}
    \label{fig:data_source}
\end{figure*}

\noindent\textbf{Dataset Visualization. }
As displayed in Figure~\ref{fig:data_source}, we compared samples from the MMC4~\cite{zhu2024multimodal} dataset and our CoMM dataset derived from various data sources. The comparison reveals: 1) MMC4 exhibits relatively poor style consistency. For the same ``duck'' entity, one image is in cartoon style while another is realistic (\cf~Figure~\ref{fig:data_source}(a)). 2) Our CoMM dataset maintains high consistency in both entity representation and style across all data sources. 3) CoMM demonstrates enhanced narrative and stylistic diversity, covering different content orientations: Instructables focuses on narrative closure, StoryBird and StoryGen emphasize cartoon-style storytelling, while WikiHow and eHow concentrate on instructional content through illustrative methods.

\begin{figure*}
    \centering
    \includegraphics[width=0.95\linewidth]{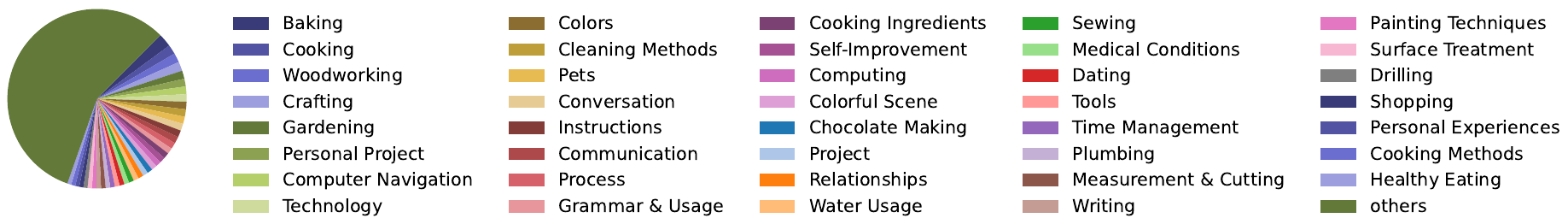}
    \caption{Topic visualization of our dataset. 'Others' contain 'Exercise', 'Drawing \& Design', 'Boating', etc., totaling 144 topics.}
    \label{fig:diversity_statics}
\end{figure*}
\noindent\textbf{Dataset Statistic.} As shown in Figure~\ref{fig:diversity_statics}, our dataset spans multiple domains, including technical fields (\eg, technology, computing), creative pursuits (\eg, crafting, painting techniques), and lifestyle areas (\eg, healthy eating, personal experiences). The distribution of topics is relatively balanced, with no single category disproportionately dominating the dataset. These statistics demonstrate the diversity of our dataset.

\section{More Experiment Studies and Analyses}
\label{sec:more_abla}

\noindent\textbf{Ablation study on the ITA.} We conduct an ablation study on various thresholds for the image-text alignment (ITA). The results are shown in Table~\ref{table:abla_ita}. We exclude data with an image-text alignment score below 4 in other experiments.

\noindent\textbf{Ablation study on the ImgS.} Table~\ref{table:abla_imgs} displays the ablation study results of various image sequence (ImgS) scores. As seen, the score of 1.0 is best for filtering.

\noindent\textbf{Ablation study on the MLLMs for dataset evaluation.} To further verify the robustness of our dataset and evaluation results against model bias, we adopt Claude-3.5-Sonnet and human for the dataset quality evaluation process. The experimental results are shown in the Table~\ref{table:mllm_data_eval}. Since Claude-3.5-Sonnet is not used for dataset construction, the evaluation results are not affected by the bias of the model used for filtering. The differences in Claude-3.5’s results among the three datasets are consistent with GPT-4o's results (\cf Table~\ref{table:comp_with_previous}), where our dataset is superior to MMC4 and OBELICS in text development, completeness, and image-text alignment.

\noindent\textbf{Scale law analysis.}
Table~\ref{table:scale_law} shows our results trained with varying percentages of our dataset, following the same settings as Table~\ref{table:perf_flamingo}. The evaluations are conducted on COCO and Flickr30K image captions in a 16-shot approach. Performance improves significantly as the data size of our dataset from 0\% to 30\%, highlighting the effectiveness of high-quality interleave data, and continues to improve gradually as data scale increases. Due to the high cost of acquiring high-quality interleaved image-text data, we cannot currently expand our dataset (with 2.28 million images) to billions. We believe that dataset quality is as important as scale.

\noindent\textbf{Human study.} To demonstrate the validity of the GPT-4o assessment, we further conduct a human study to assess the performance of the generation task in Table~\ref{table:human_study}. Given time and labor constraints, we evaluate only the most challenging task, task 4 ( Question-based Interleaved Image-Text Generation, \cf Sec.\ref{sec:task4}). Results were collected from 32 people using the same criteria as in our paper. Humans are stricter and give lower scores than GPT-4o in evaluations, but the consistent score gap trend demonstrates the feasibility of using GPT-4o for evaluations.

\noindent\textbf{Details and Additional Results Trained Using DPO~\cite{rafailov2024direct}.} We trained SEED-Llama~\cite{ge2023making} using the DPO~\cite{rafailov2024direct} algorithm with a learning rate of \(5 \times 10^{-7}\), a batch size of 16, and \(\beta\) in the DPO loss set to 0.05. The results for Task 1 are presented in our paper, and the results for the remaining tasks are shown in Table~\ref{table:more_dpo}. These findings demonstrate that the current DPO algorithm can significantly enhance the performance of text generation tasks (e.g., Task 1). However, its effectiveness in improving image generation quality remains limited, highlighting an area for further exploration and development in future research.

\noindent\textbf{Ablation Stuies among filter strategies. } We conducted ablation studies on our dataset. The results show that our high-quality data boosts performance, and our filtering strategies further enhance performance. ``Origin'' refers to using our collected data without filtering. ``ITA'' involves using LLM combined with a caption model to filter out text sequences and poorly aligned image-text data, while ``ImgS'' uses CLIP to filter image sequences. Consistent with the few-shot experiment in Table~\ref{table:filter_ablation}, we used 16 shots. 

\begin{table}[t]
\centering
\small
\setlength{\tabcolsep}{1mm}

\resizebox{0.48\textwidth}{!}{ 
\begin{tabular}{lc|ccccc}
\specialrule{0.05em}{0pt}{0pt}
\multirow{2}{*}{Methods} & \multirow{2}{*}{Size} & \multicolumn{5}{c}{T2I Sequence Generation (Task2)} \\
        &  & FID $\downarrow$ &  IS &  SSIM &  PSNR  & IRS  \\
\hline
SEED-Llama~\cite{ge2023making} & 8B & 57.96 & 8.40 & 20.53 & 7.87  & 5.27  \\
+~~~~\textbf{DPO}~\cite{rafailov2024direct} & 8B & 63.05 & 8.28 & 20.99 & 8.05 & 4.69  \\
\hdashline
SEED-Llama~\cite{ge2023making} & 14B &  66.23 & 11.06 & 20.83 & 8.12 & 6.24 \\
+~~~~\textbf{DPO}~\cite{rafailov2024direct} & 14B & 72.60 & 9.98 & 20.49 & 7.84 & 5.75  \\
\end{tabular}
}

\resizebox{0.48\textwidth}{!}{ 
\begin{tabular}{lc|cccccc}
\specialrule{0.05em}{0pt}{0pt}
\hline
\multirow{2}{*}{Methods} & \multirow{2}{*}{Size} &  \multicolumn{6}{c}{Continuation Generation (Task3)} \\
        &  &  Style  &   Entity  &    Trend & CPL. & ImgQ & IRS \\
\hline
SEED-Llama~\cite{ge2023making} & 8B &  6.28 & 5.84 & 5.72  & 6.28 & 6.55 & 2.92  \\
+~~~~\textbf{DPO}~\cite{rafailov2024direct} & 8B & 6.19 & 5.58 & 5.23 & 5.74 & 6.10 & 2.86 \\
\hdashline  
SEED-Llama~\cite{ge2023making} & 14B &  6.68 & 6.22 & 6.13 & 6.66  & 6.67 & 3.23 \\
+~~~~\textbf{DPO}~\cite{rafailov2024direct} & 14B & 6.05 & 5.54 & 5.26 &  5.97 & 6.09 & 3.11 \\
\hline
\multirow{2}{*}{Methods} & \multirow{2}{*}{Size} & \multicolumn{6}{c}{ Question-based  Generation (Task4)} \\
        &  & Style  &   Entity  &    Trend & CPL. & ImgQ & IRS \\
\hline
SEED-Llama~\cite{ge2023making} & 8B & 7.55 & 6.81 & 6.15  & 5.13 &6.36 & 1.46 \\
+~~~~\textbf{DPO}~\cite{rafailov2024direct} & 8B & 7.94 & 6.63 & 5.25 & 4.61 & 5.95 &  1.69  \\
\hdashline
SEED-Llama~\cite{ge2023making} & 14B &  7.51  & 6.61 & 6.30 & 6.13 & 6.66  & 2.50 \\
+~~~~\textbf{DPO}~\cite{rafailov2024direct} & 14B & 7.54 & 6.44 & 5.40 & 5.47 & 6.13 & 2.76 \\
\specialrule{0.05em}{0pt}{0pt}
\hline
\end{tabular}
}

\caption{Performance results of SEED-Llama~\cite{ge2023making} trained by DPO~\cite{rafailov2024direct} in our preference dataset. `CPL.'' stands for Completeness, ``ImgQ'' indicates Image Quality, ``IRS'' means Illustration Relevance Score, and ``$\downarrow$'' denotes that lower values are better.}
\label{table:more_dpo}
\end{table}

\begin{table}[t]
% \begin{wraptable}[10]{r}{7.5cm}
% \tabcolsep=0.8mm
\centering
\small
\begin{tabular}{c|cccc}
% \noalign{\hrule height 1.5pt}
\specialrule{0.05em}{0pt}{0pt}
\hline
    ITA score & 2 & 3 & 4 & 5 \\ \hline
    COCO & 103.1 & 107.1 & 109.3 & 109.1 \\ \hline
    Flickr30K & 69.3 & 70.7 & 71.5 & 70.9 \\ \hline
\specialrule{0.05em}{0pt}{0pt}
\hline
\end{tabular}
\caption{Ablation study on various image-text alignment (ITA) thresholds for data filtering. We train the model with filtered data and evaluate it using a 16-shot image caption way.}
\label{table:abla_ita}
\end{table}

\begin{table}[t]
% \begin{wraptable}[10]{r}{7.5cm}
% \tabcolsep=0.8mm
\centering
\small
\begin{tabular}{c|cccc}
% \noalign{\hrule height 1.5pt}
\specialrule{0.05em}{0pt}{0pt}
\hline
    ImgS Score & 0.5 & 1.0 & 1.5 & 2 \\ \hline
    COCO & 108.5 & 109.3 & 108.7 & 109.2 \\ \hline
    Flickr30K & 70.2 & 71.5 & 71.6 & 71.3 \\
        \hline
\specialrule{0.05em}{0pt}{0pt}
\hline
\end{tabular}
\caption{Ablation study on various our proposed image sequence (ImgS) score thresholds for data filtering. We train the model with filtered data and evaluate it using a 16-shot image caption way.}
\label{table:abla_imgs}
\end{table}
% \begin{table}[t]
% % \begin{wraptable}[10]{r}{7.5cm}
% \tabcolsep=1.5mm
% \centering
% \small
% \begin{tabular}{c|cccc}
% % \noalign{\hrule height 1.5pt}
% \specialrule{0.05em}{0pt}{0pt}
% \hline
% Models & DLP & CPL & ITA \\ \hline
% MMC4 & 4.61 & 4.76 & 4.30 \\ \hline
% OBELICS & 5.35 & 5.28 & 3.22 \\ \hline
% CoMM (\textbf{Ours}) & 7.58 & 7.12 & 8.29 \\ \hline
% \specialrule{0.05em}{0pt}{0pt}
% \hline
% \end{tabular}
% \caption{Dataset quality evaluated by Claude-3.5-Sonnet. ``DLP'' stands for Development, ``CPL'' signifies Completeness, and ``ITA'' represents Image-Text Alignment. Each dataset is randomly sampled with 5000 cases for evaluation.}
% \label{table:mllm_data_eval}
% \end{table}

\begin{table}[t]
\tabcolsep=1.5mm
\centering
\small
\begin{tabular}{c|ccc}
\specialrule{0.05em}{0pt}{0pt}
\hline
\multirow{2}{*}{Models} & DLP & CPL & ITA \\
 & CL / HM & CL / HM & CL / HM \\
\hline
MMC4 & 4.61 / 6.20 & 4.76 / 6.48 & 4.30 / 5.79 \\
OBELICS & 5.35 / 5.58 & 5.28 / 5.60 & 3.22 / 4.87 \\
CoMM (\textbf{Ours}) & \textbf{7.58} / \textbf{8.32} & \textbf{7.12} / \textbf{8.57} & \textbf{8.29} / \textbf{8.56} \\
\specialrule{0.05em}{0pt}{0pt}
\hline
\end{tabular}
\caption{Dataset quality evaluated by Claude-3.5-Sonnet (CL) and Human (HM). ``DLP'' stands for Development, ``CPL'' signifies Completeness, and ``ITA'' represents Image-Text Alignment. For CL, Each dataset is randomly sampled with 5000 cases for evaluation. For HM, a total of 570 feedback responses were collected from 19 persons, each evaluating 30 documents.}
\label{table:mllm_data_eval}
\end{table}

\begin{table*}[t]
% \begin{wraptable}[10]{r}{7.5cm}
\tabcolsep=1.5mm
\centering
\small
\begin{tabular}{c|cccc}
% \noalign{\hrule height 1.5pt}
\specialrule{0.05em}{0pt}{0pt}
\hline
Percent (Image Number) & \multicolumn{1}{l}{\textbf{0\% (Baseline)}} & \multicolumn{1}{l}{\textbf{30\% (0.68 M)}} & \multicolumn{1}{l}{\textbf{60\% (1.37 M)}} & \multicolumn{1}{l}{\textbf{100\% (2.28 M)}} \\
\hline
COCO  & 98.8  & 105.3 & 106.9 & 109.3 \\
Flickr30K & 62.8  & 68.7  & 70.8  & 71.5 \\
\specialrule{0.05em}{0pt}{0pt}
\hline
\end{tabular}
\caption{Scale law analysis.}
\label{table:scale_law}
\end{table*}
\begin{table}[t]
% \begin{wraptable}[10]{r}{7.5cm}
\tabcolsep=1.0mm
\centering
\small
\begin{tabular}{lc|ccccc}
% \noalign{\hrule height 1.5pt}
\specialrule{0.05em}{0pt}{0pt}
\hline
  Models & Size  & \multicolumn{1}{c}{\textbf{Style}} & \multicolumn{1}{c}{\textbf{Entity}} & \multicolumn{1}{c}{\textbf{Trend}} & \multicolumn{1}{c}{\textbf{CPL}} & \multicolumn{1}{c}{\textbf{ImgQ}} \\
\hline
    MiniGPT-5~\cite{zheng2023minigpt} & 7B    & 6.39  & 5.92  & 5.84  & 6.67  & 6.16 \\
    SEED-Llama~\cite{ge2023making} & 8B    & 6.79  & 6.48  & 5.48  & 6.68  & 6.24 \\
    SEED-Llama~\cite{ge2023making} & 14B   & 6.83  & 6.37  & 5.54  & 6.37  & 6.33 \\
    Emu2~\cite{sun2023generative}  & 33B   & 7.27  & 6.56  & 5.47  & 6.79  & 6.03 \\
\specialrule{0.05em}{0pt}{0pt}
\hline
\end{tabular}
\caption{Human study on question-based interleaved image-text generation task.}
\label{table:human_study}
\end{table}
\begin{table}
\tabcolsep=1.5mm
\centering
\small
\vspace{-1.3em}
% \setlength\tabcolsep{1pt}
% \hspace{-2.6em}
\begin{tabular}{ccc|ccc}
\specialrule{0.05em}{0pt}{0pt}
\hline
Origin & ITA & ImgS & COCO & Flickr30k & TextVQA \\ \hline
            &           &               & 98.8 & 62.8      & 27.3    \\
\checkmark  &           &               & 105.7 & 67.4     & 31.5    \\
\checkmark  & \checkmark &               & 108.5 & 70.2    & 34.2    \\
\checkmark  &           & \checkmark    & 107.7 & 68.3    & 33.6    \\
\checkmark  & \checkmark & \checkmark   & 109.3 & 71.5    & 35.8    \\ 
\hline
\specialrule{0.05em}{0pt}{0pt}
\end{tabular}
\caption{Ablation studies among filter strategies. }
\label{table:filter_ablation}
\end{table}
\begin{table}[h]
\centering
\small
\setlength{\tabcolsep}{2mm} 
\begin{tabular}{l|cccccc}
\specialrule{0.05em}{0pt}{0pt}
\hline
Dataset & Style & Entity & Trend & CPL & ImgQ & IRS \\
\hline
 & \multicolumn{6}{c}{Continuation Generation (Task 3)}  \\
{MMC4} & 5.22 & 4.90 & 4.47 & 4.45 & 5.5 & 1.32 \\
{OBELICS} & 4.67 & 4.10 & 3.83 & 3.58 & 5.26 & 0.96 \\
{\textbf{CoMM}} & \textbf{6.68} & \textbf{6.22} & \textbf{6.13} & \textbf{6.66} & \textbf{6.67} & \textbf{3.23} \\
\hline
 & \multicolumn{6}{c}{Question-based Generation (Task 4)} \\
{MMC4} & 5.78 & 4.71 & 3.58 & 3.04 & 4.31 & 1.31 \\
{OBELICS} & 3.25 & 2.65 & 1.82 & 1.84 & 4.59 & 1.14 \\
{\textbf{CoMM}} & \textbf{7.51} & \textbf{6.61} & \textbf{6.30} & \textbf{6.13} & \textbf{6.66} & \textbf{2.50} \\
\specialrule{0.05em}{0pt}{0pt}
\hline
\end{tabular}
\caption{Performance comparison among three datasets on Task 3 \& 4. We train different datasets on SEED-Llama-14B. We used LLM to generate titles for MMC4 and OBELICS, which served as pseudo-label questions for training.}
\label{table:vertical_combined_tasks}
\end{table}

\section{Ethical Discussion and License}\label{sec:ethic_impact}

\textbf{Ethical Discussion. }
Collecting data from online sources comes with the risk of encountering content that may not be suitable for all audiences. Fortunately, this risk is minimized in our case because we focus on high-quality data, such as instructional steps and visual stories. Besides, these specific websites have their review/
editorial processes, which significantly improve data quality and reduce potential hazards. For example, WikiHow~\cite{wikihow} claims that "the average WikiHow article has been edited by 23 people and reviewed by 16 people". To further ensure the integrity of our dataset, we perform a rigorous screening process to filter out any NSFW content (as mentioned in Sec~\textcolor{red}{3.1}), trying to maintain a clean, and reliable dataset suitable for all users.

\noindent\textbf{License and Author Statement. }
We release this dataset under a CC-BY license and Terms of Use that require disclosure when used for model training. This license does not override the original content licenses; all use must comply with the original licenses and data subjects' rights. We clarify the user's responsibilities and liabilities here. While we've tried our best to ensure data accuracy and legality, we cannot guarantee absolute correctness. We assume no liability for rights violations, including but not limited to copyright, privacy issues, or misuse of sensitive information. 

By using this dataset, you accept full responsibility for legal or other consequences. You agree to adhere to all relevant laws, regulations, and ethical guidelines. Accessing or using this dataset signifies your acceptance of this statement and the CC-BY license terms. Disagreement with these terms means you are not authorized to use the dataset.

\section{Filter Strategy and Evaluation Details
% Prompt Details
}\label{sec:prompt_details}
\subsection{Data Quality Filter Prompt. }
Below is the prompt for ensuring data quality in text and image-text alignment. When using GPT-4o~\cite{openai2024gpt4o}, which can see images directly, we input the original images directly. However, when using Llama3~\cite{meta2024llama3}, which cannot see images, we first employ CogVLM~\cite{wang2023cogvlm} to convert the image into a detailed caption, then input it in the ``<IMAGE>image description</IMAGE>'' format.

\begin{lstlisting}[
    basicstyle=\ttfamily\small, % Basic font style and size
    breaklines=true, % Enable line breaking
    numbers=none, % Disable line numbers
    backgroundcolor=\color{yellow!10}, % Background color
    frame=single, % Frame around the code
    rulecolor=\color{yellow!50!black}, % Frame color
    frameround=tttt, % Rounded corners
    showstringspaces=false, % Do not show spaces as special characters
    keywordstyle=\color{black},
    belowskip=0pt 
]
You are a master of multi-modal evaluation. Your task is to evaluate the quality of a docs that contains images and text. Images will be presented in the format <IMAGE>image description</IMAGE>. The textual content will be presented as plain text. Evaluate the following criteria:
1. Development: Assess the coherence and logical flow of the data. Only the most logically consistent and well-integrated contexts should receive high scores.
2. Completeness: Check if the content provides a comprehensive and detailed overview of the topic. Full scores should only be given for thorough and exhaustive coverage.
3. Interleaving of Images and Text: Ensure that the images and text are perfectly aligned. Discrepancies or inconsistencies should result in significant deductions.

Scores should range widely to highlight exceptional quality or notable deficiencies. Each criterion should be evaluated and concluded with a score on a scale from 0 to 10, where 0-2 indicates major deficiencies and 8-10 indicates exemplary performance. Structure your response as follows:
<Development>
  <Problem>Brief description of any issues</Problem>
  <Score>Numerical rating</Score>
</Development>
<Completeness>
  <Problem>Brief description of any gaps</Problem>
  <Score>Numerical rating</Score>
</Completeness>
<Image-Text Interleaving>
  <Problem>Brief description of any discrepancies</Problem>
  <Score>Numerical rating</Score>
</Image-Text Interleaving>

Emphasize the identification of particularly strong or weak points in the Problem section. This feedback will guide you to adjust scores to be more polarized, reflecting a clear distinction between high and low quality.

Data to Review:
<data>
{}
</data>
\end{lstlisting}
% \end{tcolorbox}

\subsection{Evaluation Prompt for Interleaved Generation Content. }
We explain the motivation and detailed prompt design of the GPT-4o evaluation here.

\textbf{Document Completeness, Image Sequence Coherence and Image Quality.} 
Here are the prompts for evaluating the document completeness, image sequence coherence, and image quality.

\begin{lstlisting}[
    basicstyle=\ttfamily\small, % Basic font style and size
    breaklines=true, % Enable line breaking
    numbers=none, % Disable line numbers
    backgroundcolor=\color{yellow!10}, % Background color
    frame=single, % Frame around the code
    rulecolor=\color{yellow!50!black}, % Frame color
    frameround=tttt, % Rounded corners
    showstringspaces=false, % Do not show spaces as special characters
    keywordstyle=\color{black},
    belowskip=0pt 
]
We are evaluating the results of a model designed for generating interleaved image-text documents. The model's input, starting with "INPUT:", can either be the beginning of a text-image interleaved document or a specified topic. Its output, starting with "OUTPUT:", will then be either a continuation of the document or content generated based on the given topic. The image with the index i will be enclosed by the symbols "<Img_i>" and "</Img_i>". The images are numbered sequentially from 0 to N (including the input images).
As an expert in multimodal evaluation, your task is to assess the quality of the output that includes both images and text. The images are numbered sequentially from 1 to n (include the input images). Use the guidelines below to assign a final score.

Scoring Guidelines:
- 0-3: Major deficiencies, misalignment, or inconsistency
- 4-7: Minor gaps, misalignment, or inconsistency
- 8-10: Complete and thorough alignment, strong consistency

Scoring Criteria:

1. Image Coherence:
   - Evaluate the consistency of style and entity between the output images. Assess whether the trend shown by the image sequence aligns with the text. Finally, an overall consistency score will be assigned to the image sequence.

2. Completeness:
   - Summarize the output document's topic and evaluate how thorough and comprehensive the output content is.
       Evaluate Thoroughness and Comprehensiveness:
         Is the text content complete? Is there anything missing?
         Is the image content complete? Are any images missing?
         Do the images and text fully support each other? Is there any missing image or text?

3. Image Quality:
   - Evaluate the quality of the output images based on the following aspects:
       Realism: Determine whether the image resembles a real scene or object and identify any signs of artificial model synthesis.
       Completeness: Check if the objects in the image are fully intact, without any noticeable missing parts, truncation, or damage.
       Clarity: Determine if the details are sufficient and if the image is free of blurriness or out-of-focus areas.
       Composition balance: Evaluate the aesthetic quality and balance of the image composition, ensuring that the main subjects are well-framed and the composition is visually pleasing.

Assume the index of the first image in the output is K.
JSON Output Structure: 
{
    "scores": {
        "Image_Coherence": {
            "pair_scores": {
                "image_K_and_K+1": {
                    "style_consistency": 0-10,
                    "entity_consistency": 0-10,
                    "justification": "Brief explanation of any gap"
                },
                "image_K+1_and_K+2": {
                    "style_consistency": 0-10,
                    "entity_consistency": 0-10,
                    "justification": "Brief explanation of any gap"
                }
                // Continue for remaining pairs...
            },
            "overall_score": {
                "style_consistency": 0-10,
                "entity_consistency": 0-10,
                "trend_consistency": 0-10,
                "overall_consistency": 0-10,
                "justification": "Brief explanation of overall consistency"
            }
        },
        "Completeness": {
            "Summarize": "brief summary",
            "Justification": "brief justification of any issue",
            "Score": 0-10
            },
        "Image_Quality":{
            "Score": 0-10,
            "Justification": "brief justification of any deficiencies in image quality",
        }
    }
}

Data to Review: 
\end{lstlisting}

\textbf{Illustration Relevance Score.}
In our evaluation tests, we found that GPT-4o has a good understanding of images, but the text context will seriously influence this understanding. For example, when only images from a step-by-step instruction document are input, GPT-4o can accurately describe both the correctly ordered and reversed-ordered image contents. However, when the interleaved texts and images are input together, GPT-4o tends to produce similar descriptions for both the correctly ordered and reversed-ordered images, which is incorrect.

Thus we design an evaluation process similar to human document creation to mitigate this limitation. We first generate the required image content based on the text context and then evaluate the consistency between the image descriptions and the corresponding images.
Specifically, this process involves two model invocations, the following are the prompts for the first invocation:

\begin{lstlisting}[
    basicstyle=\ttfamily\small, % Basic font style and size
    breaklines=true, % Enable line breaking
    numbers=none, % Disable line numbers
    backgroundcolor=\color{yellow!10}, % Background color
    frame=single, % Frame around the code
    rulecolor=\color{yellow!50!black}, % Frame color
    frameround=tttt, % Rounded corners
    showstringspaces=false, % Do not show spaces as special characters
    keywordstyle=\color{black},
    belowskip=0pt 
]
We are evaluating the results of a model designed for generating interleaved image-text documents.  The model's input, starting with "INPUT:", can either be the beginning of a text-image interleaved document or a specified topic. Its output, starting with "OUTPUT:", will then be either a continuation of the document or content generated based on the given topic.  The image with the index i will be enclosed by the symbols "<Img_i>" and "</Img_i>". The images are numbered sequentially from 0 to N (include the input images). Now we hide the output's images while preserving the "<Img_i></Img_i>". As an expert in multimodal evaluation, you are responsible for predicting the removed image's content based on the input and the output text context.

Tasks:
1. Predict Each Image's Content:
    For each image content prediction, predict the most probable and suitable image content based on the input and text context in the output. The description should consider the illustration needs (What should the image illustrate to complement its surrounding text context?), content description (Provide a detailed description of what the image should contain.), and context coherence (Ensure that the final narrative flows well and forms a complete, coherent document.).

Assume the index of the first removed image in the output is K.
JSON Output Structure: 
{
    "Tasks": {
        "Create an Interleaved Text-Image Document": {
            "Content of Image K": "predicted content of image K",
            "Content of Image K+1": "predicted content of image K+1",
            ...
            "Content of Image N": "predicted content of image N",
        }
    }
}

Data to Review: 
\end{lstlisting}

After the first invocation, we reorganize the output descriptions with corresponding images and start the second model invocation with the following prompts:

\begin{lstlisting}[
    basicstyle=\ttfamily\small, % Basic font style and size
    breaklines=true, % Enable line breaking
    numbers=none, % Disable line numbers
    backgroundcolor=\color{yellow!10}, % Background color
    frame=single, % Frame around the code
    rulecolor=\color{yellow!50!black}, % Frame color
    frameround=tttt, % Rounded corners
    showstringspaces=false, % Do not show spaces as special characters
    keywordstyle=\color{black},
    belowskip=0pt 
]
As an expert in image description evaluation, your job is to assess the consistency between two sets of images and their corresponding descriptions. Use the criteria below to assign a final score.
The input will be formatted as description-image pairs like <Description_i> image description </Description_i> <Img_i> image </Img_i>. Note that sometimes one of the descriptions and the image is missing, just score that input data as 0!

Scoring Guidelines:
  0-3: Major deficiencies/misalignment/inconsistency,
  4-7: Minor gaps/misalignment/inconsistency,
  8-10: Complete and thorough alignment, strong consistency.

Scoring Criteria:

1. Consistency:
    - Task: Evaluate the consistency between each image and its corresponding description.
JSON Output Structure: 
{
  "Consistency": {
      "image_1_score": 0-10,
      "image_2_score": 0-10,
      ...
      "image_n_score": 0-10,
      "overall_score": 0-10,
      "Justification": "Brief justification of any issue identified"
  }
}
    
Data to Review:
\end{lstlisting}

The output overall score is the IRS.

\section{Model Training Detail }\label{sec:model_training}
% \textcolor{red}{Training parameter and GPUs }

% \begin{itemize}
\textbf{MiniGPT-5}~\cite{zheng2023minigpt} combines the Stable Diffusion with LLMs through ``generative vokens''. This model adopts a two-stage training strategy tailored for description-free multimodal generation. Initially, it focuses on extracting high-quality text-aligned visual features. In the subsequent stage, it ensures optimal coordination between visual and textual prompts, significantly enhancing its ability to generate coherent multimodal content. 

\textit{Training Settings. } We train MiniGPT-5 using 8 A100-80G GPUs, fine-tuning the parameters of the LoRA~\cite{hu2021lora} layers (the rank is 32) in the LLM backbone and the Feature Mapper for output visual tokens. The learning rate is set to 5e-5 for the LoRA layers and 5e-4 for the other trainable parameters, with a total of 5 training epochs. All other settings follow those of MiniGPT-5.

\textbf{SEED-Llama}~\cite{ge2023making} equips the pre-trained LLM~\cite{touvron2023llama} with a VQ-based image tokenizer (SEED), which processes images into discrete tokens. This tokenizer utilizes a 1D causal dependency to align visual tokens with the autoregressive nature of LLMs, enhancing semantic coherence between text and images. Enhanced by extensive multimodal pretraining and fine-tuning under a next-word-prediction objective, SEED-Llama excels in handling both comprehension and generation tasks within a unified multimodal framework.

\textit{Training Settings.} We train SEED-Llama-8B and SEED-Llama-14B using 8 A100-80G GPUs. Only the parameters of the LoRA~\cite{hu2021lora} layers (with a rank of 16) in the LLM backbone are fine-tuned. The learning rate is set to 1e-5, and the training consists of 10,000 steps.

\textbf{Emu2}~\cite{sun2023generative}
is a generative multimodal model, trained on large-scale multimodal sequences with a unified autoregressive objective. This model showcases significant capabilities in multimodal in-context learning, adept at complex tasks that require on-the-fly reasoning, such as visual prompting and object-grounded generation. 
% \end{itemize} 

\textit{Training Settings. } Emu2 is trained using 16 A100-80G GPUs. We fine-tune the parameters in the linear projection layer for input and output visual embeddings, as well as the LoRA~\cite{hu2021lora} layers (with a rank of 32) within the LLM backbone. The learning rate is set to 5e-5, and the training lasts for 5 epochs.

\section{More Generation Visualization }
\label{sec:more_vis}

\begin{figure*}
    \centering
    \includegraphics[width=0.8\linewidth]{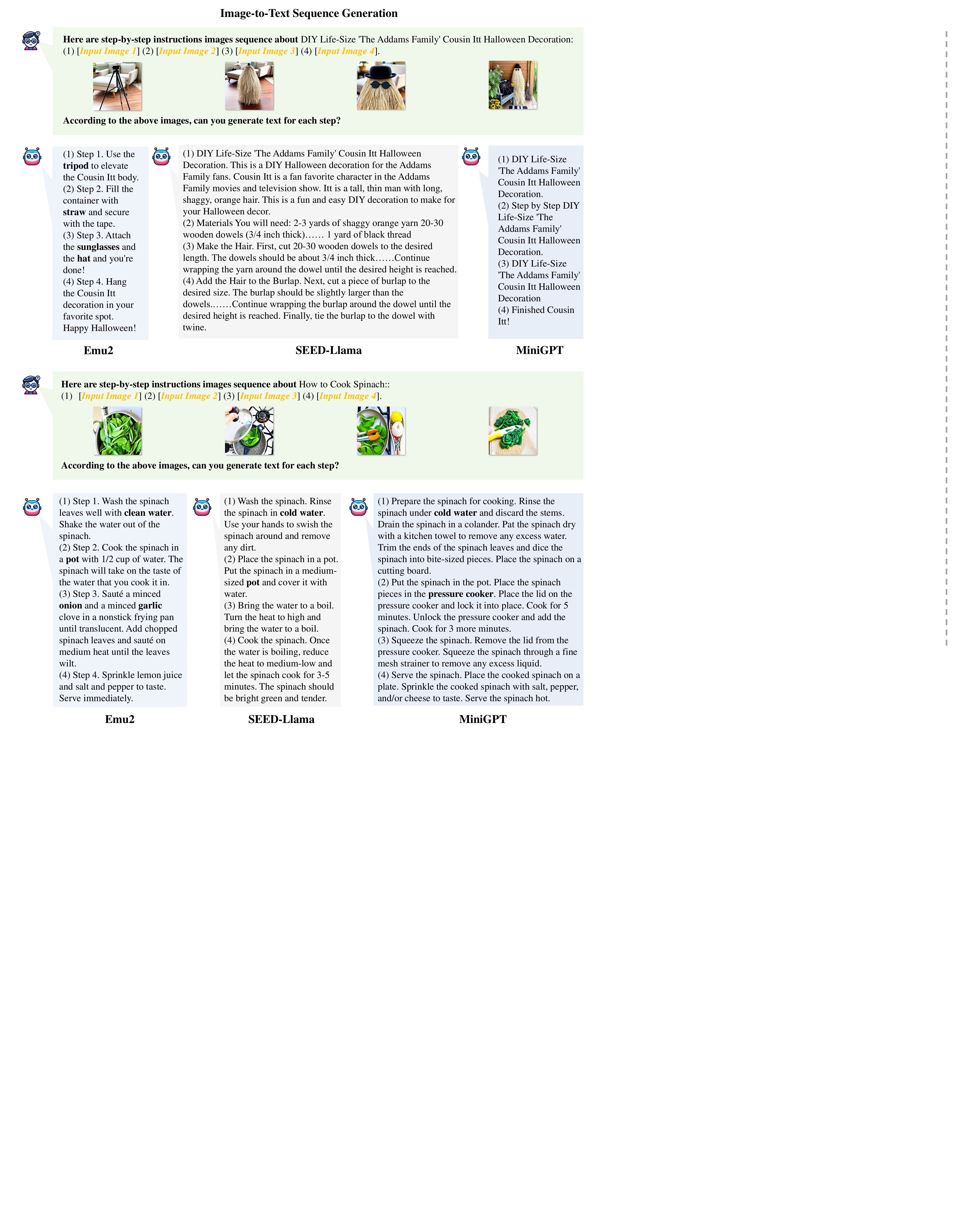}
    \caption{Visualization of image-to-text sequence generation from Emu2~\cite{sun2023generative}, SEED-Llama~\cite{ge2023making}, and MiniGPT-5~\cite{zheng2023minigpt}, separately.}
    \label{fig:task1}
\end{figure*}

\begin{figure*}
    \centering
    \includegraphics[width=0.8\linewidth]{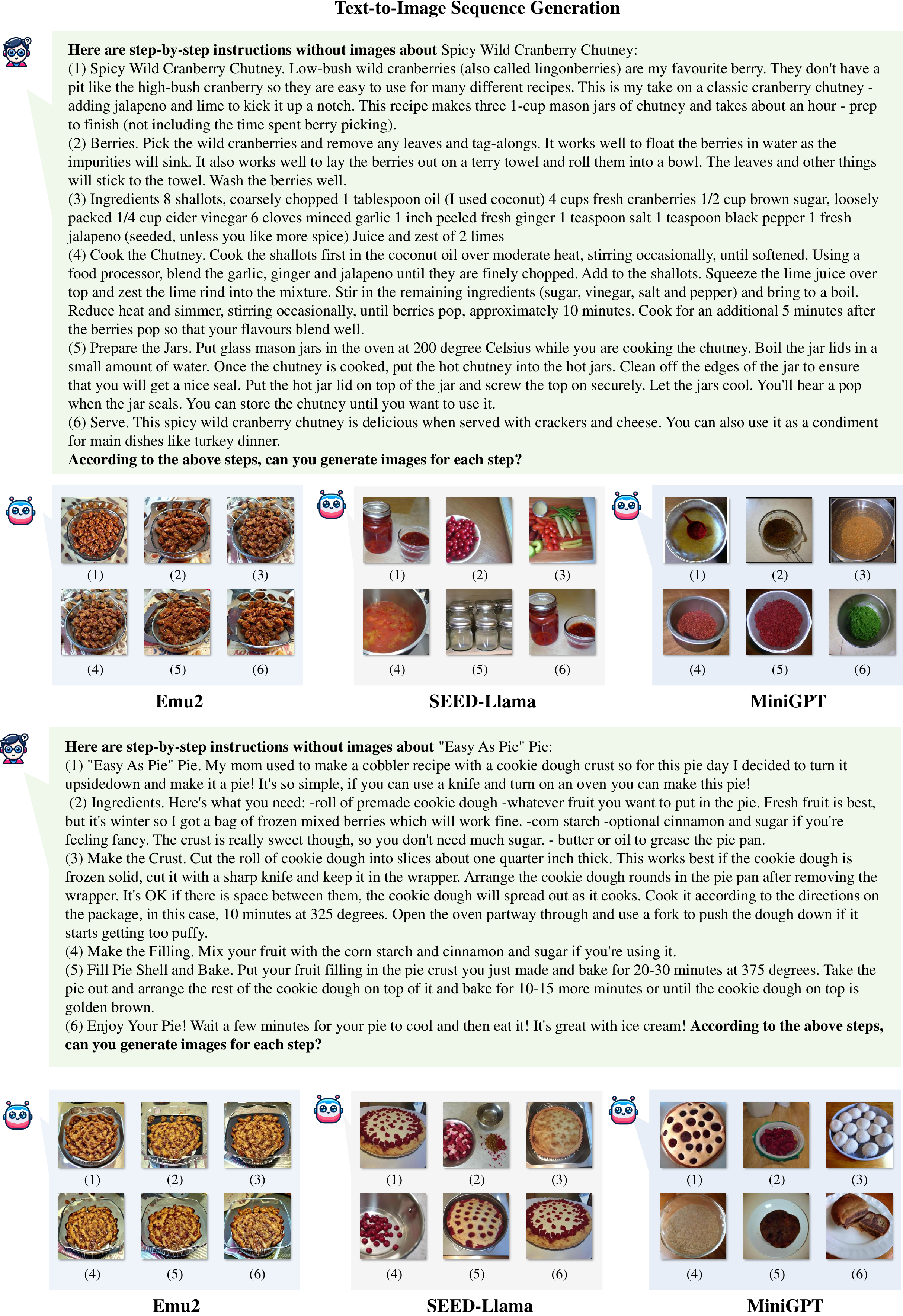}
    \caption{Visualization of text-to-image sequence generation from Emu2~\cite{sun2023generative}, SEED-Llama~\cite{ge2023making}, and MiniGPT-5~\cite{zheng2023minigpt}, separately.}
    \label{fig:task2}
\end{figure*}

\begin{figure*}
    \centering
    \includegraphics[width=0.8\linewidth]{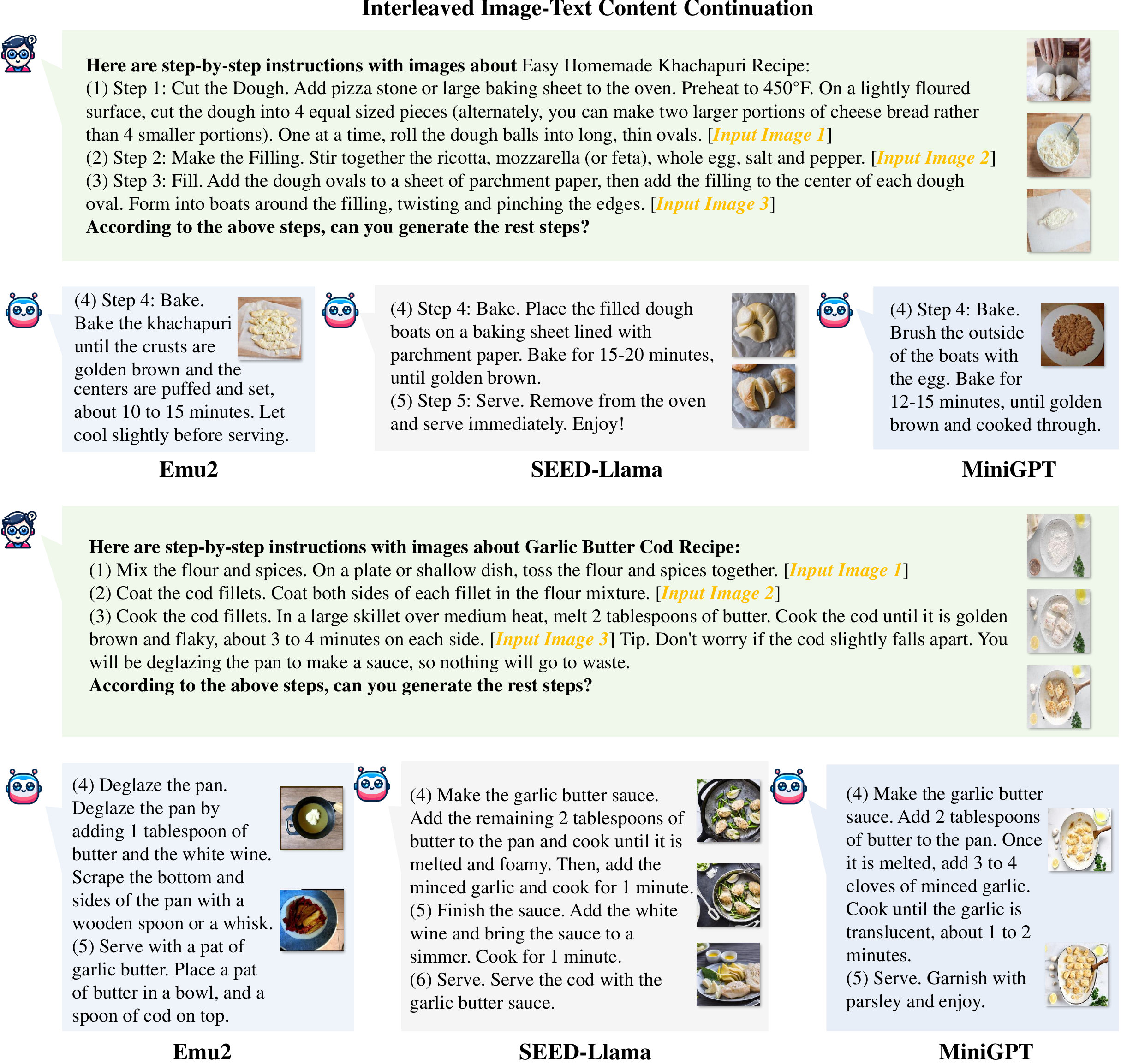}
    \caption{Visualization of interleaved image-text content continuation from Emu2~\cite{sun2023generative}, SEED-Llama~\cite{ge2023making}, and MiniGPT-5~\cite{zheng2023minigpt}, separately.}
    \label{fig:task3}
\end{figure*}

\begin{figure*}
    \centering
    \includegraphics[width=0.8\linewidth]{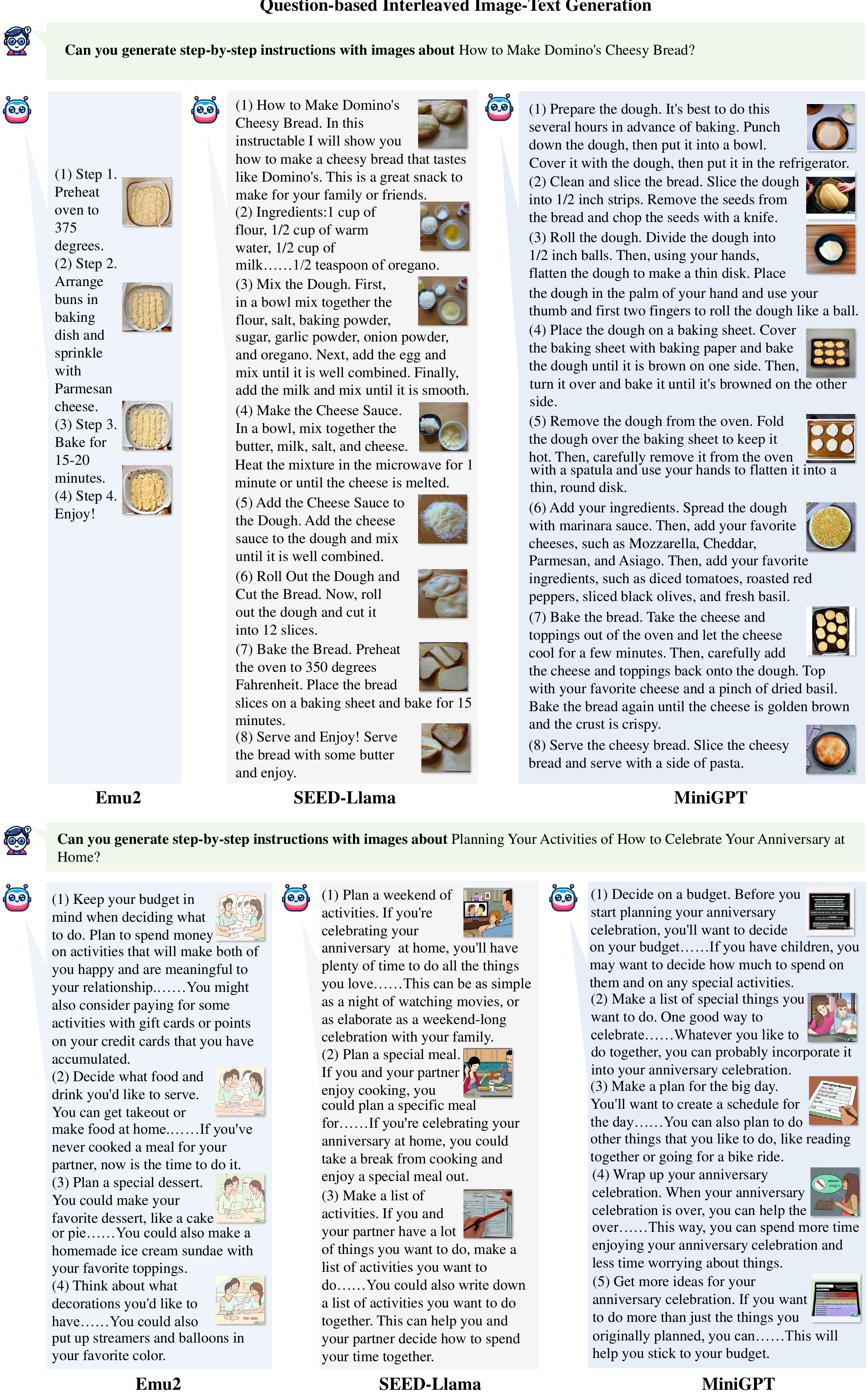}
    \caption{Visualization of question-based interleaved image-Text generation from Emu2~\cite{sun2023generative}, SEED-Llama~\cite{ge2023making}, and MiniGPT-5~\cite{zheng2023minigpt}, separately.}
    \label{fig:task4}
\end{figure*}

\textbf{Qualitative Analysis of Interleaved Generation}
We visualized the results of three baseline models (Emu2~\cite{sun2023generative}, SEED-Llama~\cite{ge2023making}, and MiniGPT-5~\cite{zheng2023minigpt}) across four interleaved generation tasks: image-to-text sequence generation (\cf~Figure~\ref{fig:task1}), text-to-image sequence generation (\cf~Figure~\ref{fig:task2}), interleaved image-text content continuation (\cf~Figure~\ref{fig:task3}), and question-based interleaved image-text Generation (\cf~Figure~\ref{fig:task4}). From the results, we can observe that: 1) For the single textual modality generation, the Emu2 model can more accurately describe entities (\eg, tripod, straw, and sunglasses in Figure~\ref{fig:task1}) appearing in images, producing more coherent and concise text descriptions. 2) For the single visual modality generation, SEED-Llama can generate images that are highly aligned with texts and and exhibit consistent style, as exemplified by the uniformity in shape and decoration of the pie depicted in Figure~\ref{fig:task2}. 3) For interleaved image-text generation, 
SEED-Llama also exhibits more coherent and stylistically consistent image-text outputs. Conversely, the Emu2 model generates images with excessive uniformity, which detracts from their developmental progression. While MiniGPT displays some developmental aspects, it suffers from inconsistencies in style.

\begin{figure*}
    \centering
    \includegraphics[width=0.8\linewidth]{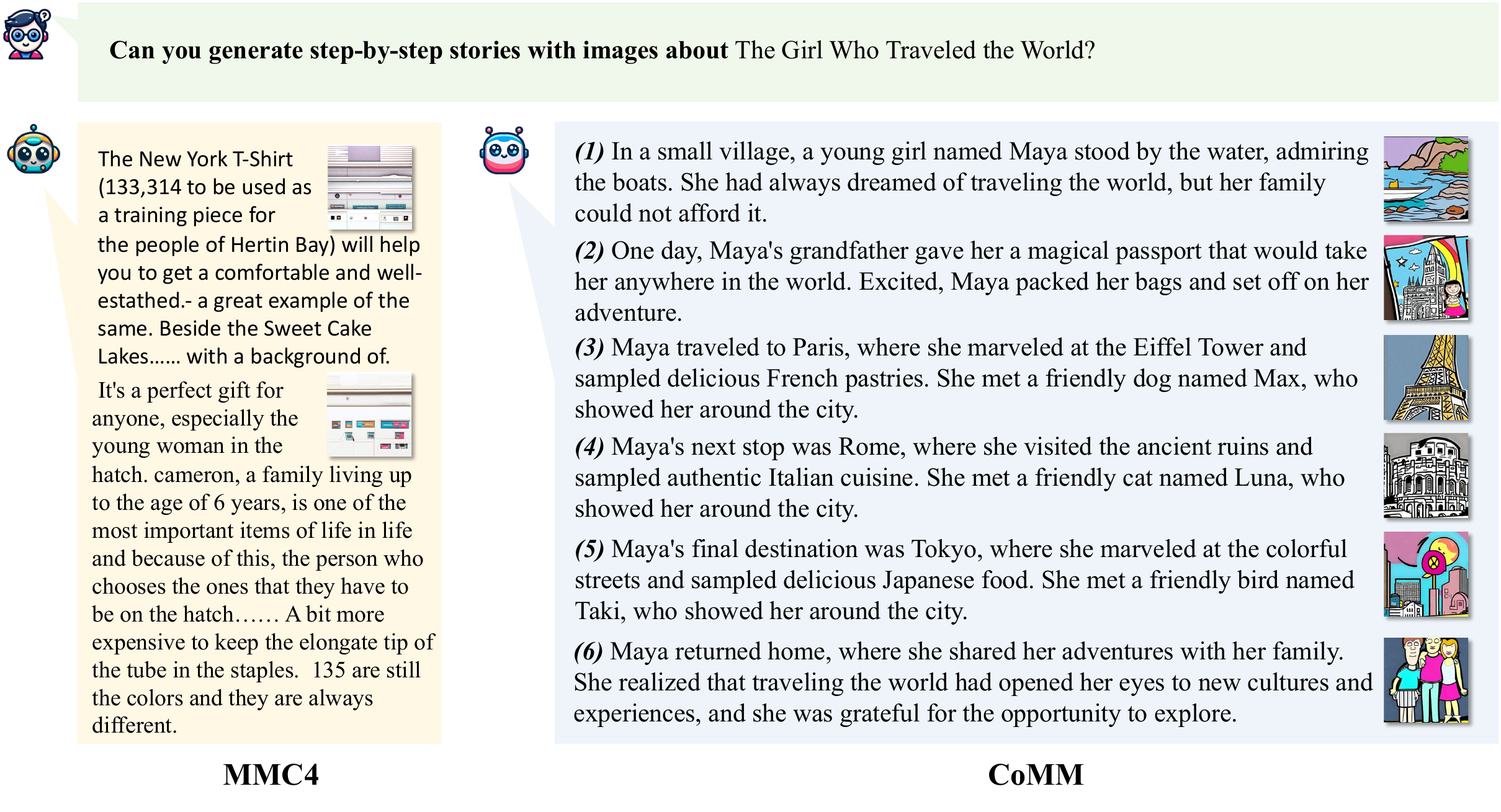}
    \caption{Comparison of storytelling visualization between SEED-Llama~\cite{ge2023making} trained on MMC4 and CoMM (ours).}
    \label{fig:story}
\end{figure*}

In addition, to demonstrate that MLLMs can facilitate a variety of applications with the interleaved multimodal generation ability, we visualize the SEED-Llama’s generation results of visual storytelling with MMC4 and our CoMM, respectively (\cf Figure~\ref{fig:story}). As seen, compared to MMC4, our CoMM enables the SEED-Llama to generate stories with more consistent style and coherent content.

\end{document}